\documentclass[sigconf, nonacm]{acmart}
\usepackage{hhline}
\usepackage[noEnd=true]{algpseudocodex}
\usepackage{algorithm}
\usepackage{pgfplots,tikz}
\usepgfplotslibrary{statistics}
 \usepackage{array,multirow,graphicx}
\usetikzlibrary{matrix}
\usepackage[normalem]{ulem} 
\pgfplotsset{compat=1.9}
\usepackage{subcaption}
\usepackage{graphicx}
\usepackage{multicol}
\usepackage{mathtools}
\usepackage{float}
\usepackage{extdash}
\DeclareMathOperator*{\concat}{%
    \mathchoice%
        {\Big\Vert}%
        {\big\Vert}%
        {\Vert}%
        {\Vert}%
}

\begin{document}
\title{LightTS: Lightweight Time Series Classification with Adaptive Ensemble Distillation---Extended Version} 

\author{\texorpdfstring{David Campos$^1$, Miao Zhang$^{1,2}$, Bin Yang$^{1,3}$, Tung Kieu$^1$, Chenjuan Guo$^{1,3}$, and Christian S. Jensen$^1$}{David Campos, Miao Zhang, Bin Yang, Tung Kieu, Chenjuan Guo, and Christian S. Jensen}}

\affiliation{%
  \institution{$^1$Aalborg University, Denmark $^2$Harbin Institute of Technology, Shenzhen, China $^3$East China Normal University, China}
  \country{$^1$\{dgcc,tungkvt,csj\}@cs.aau.dk $^2$zhangmiao@hit.edu.cn $^3$\{byang, cjguo\}@dase.ecnu.edu.cn}
}

\begin{abstract}

Due to the sweeping digitalization of processes, increasingly vast amounts of time series data are being produced. Accurate classification of such time series facilitates decision making in multiple domains. State-of-the-art classification accuracy is often achieved by ensemble learning where results are synthesized from multiple base models. This characteristic implies that ensemble learning needs substantial computing resources, preventing their use in resource-limited environments, such as in edge devices. To extend the applicability of ensemble learning, we propose the \texttt{LightTS} framework that compresses large ensembles into lightweight models while ensuring competitive accuracy. First, we propose  adaptive ensemble distillation that assigns adaptive weights to different base models such that their varying classification capabilities contribute purposefully to the training of the lightweight model. Second, we propose means of identifying Pareto optimal settings w.r.t. model accuracy and model size, thus enabling users with a space budget to select the most accurate lightweight model. We report on experiments using 128 real-world time series sets and different types of base models that justify key decisions in the design of \texttt{LightTS} and provide evidence that \texttt{LightTS} is able to outperform competitors.

\end{abstract}
\maketitle
\pagestyle{plain}

\section{Introduction} \label{sec:introduction}

With the instrumentation of processes in a broad range of settings, time series data is increasingly prevalent. As a result, the ability to analyze and create value from time series data is increasingly important. For example, accurate classification of time series is important in applications related to health, in industrial automation, in web services, and in cybersecurity~\cite{FawazFWIM19,AcharyaFOHTA17a,WangCHPH19}.

\begin{figure*}
\vspace{-3.09cm}
\centering
\begin{subfigure}{0.28\textwidth}
    \begin{minipage}[c][11cm][t]{\textwidth}
      \vspace*{3.06cm}
      \includegraphics[width=5cm]{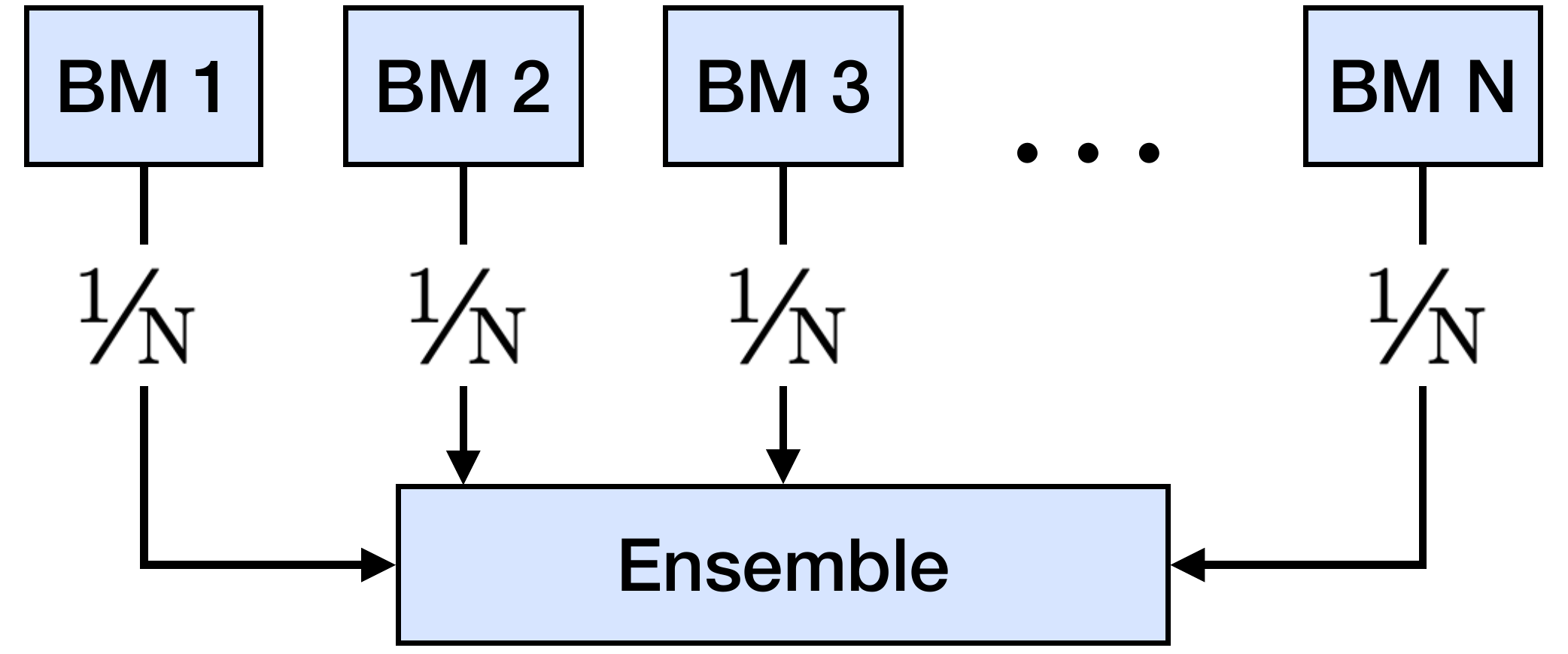}
      \caption{Ensemble classifier.}
      \label{subfig:ensemble}
      \includegraphics[width=5cm]{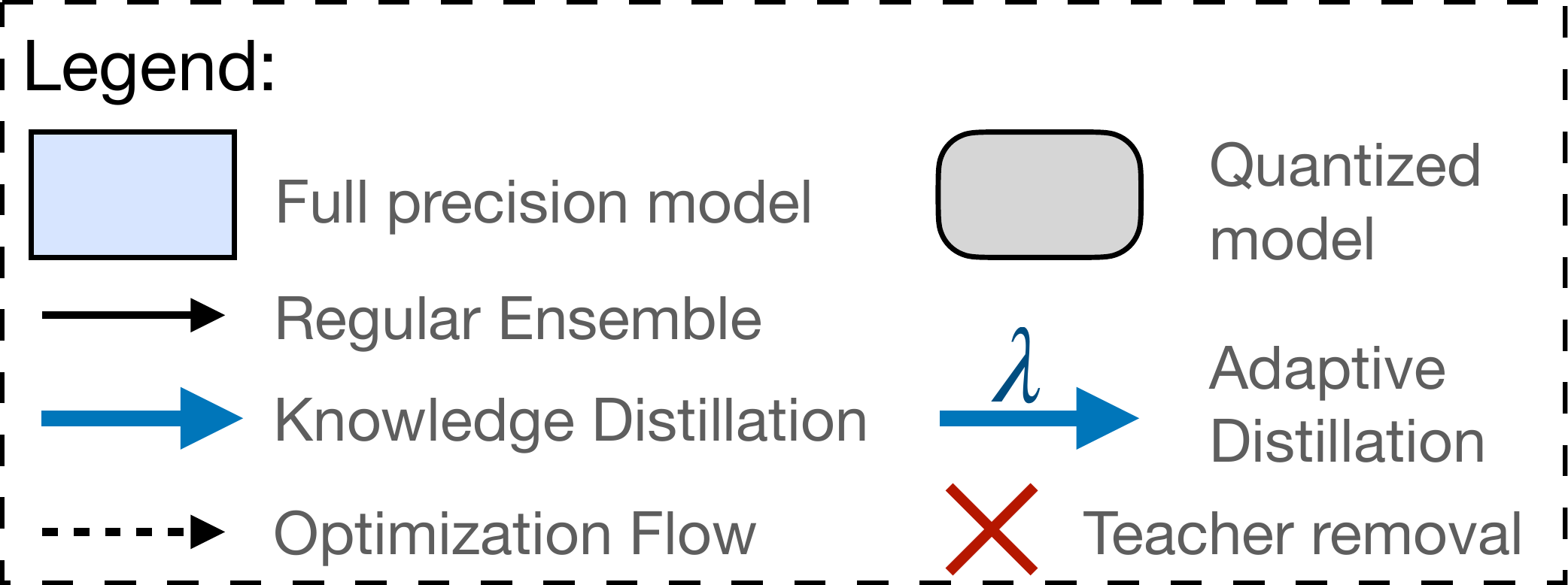}
    \end{minipage}
\end{subfigure}
\hfill
\begin{subfigure}{0.29\textwidth}
    \includegraphics[width=\textwidth]{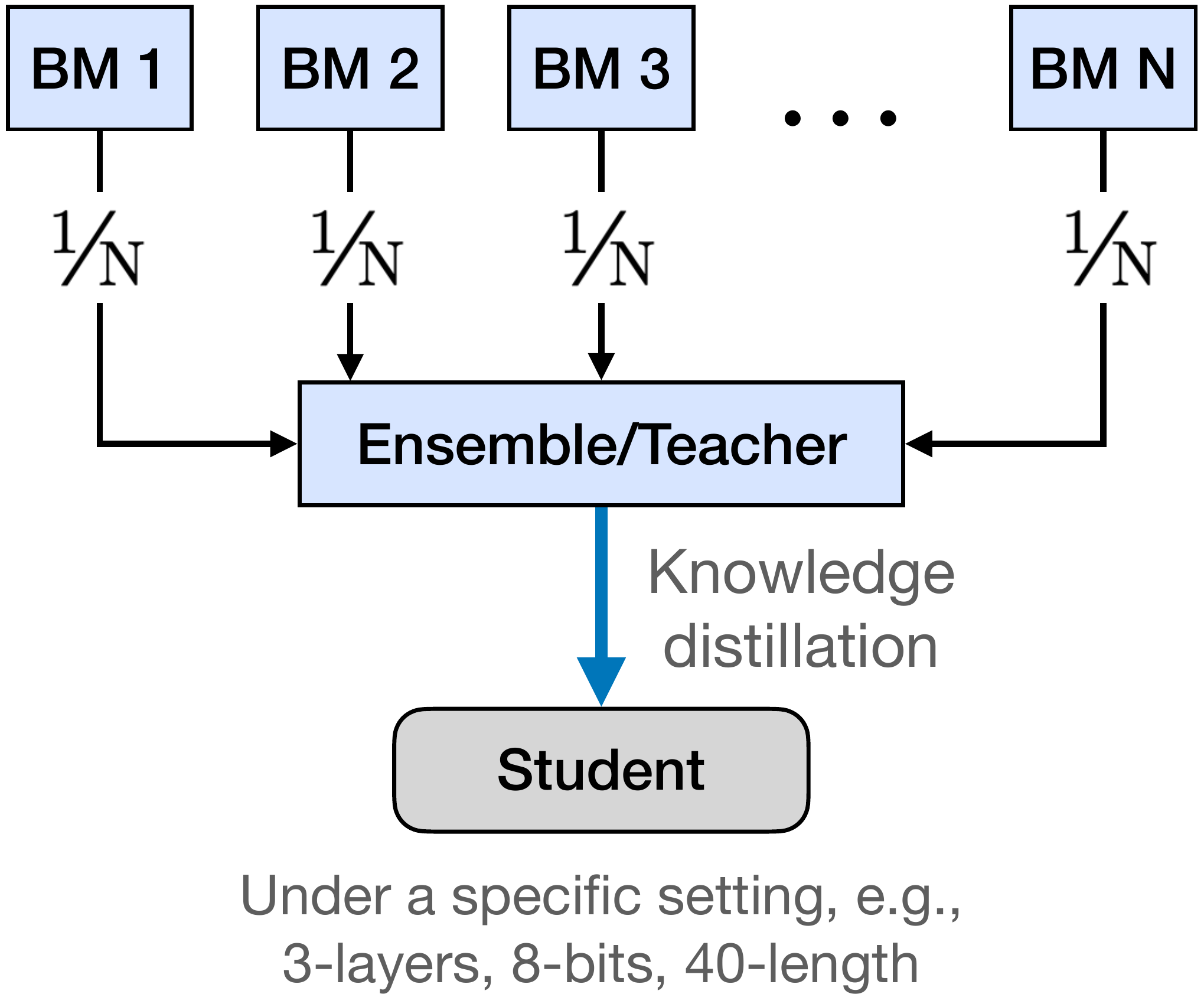}
    \caption{Knowledge distillation.}
    \label{subfig:knowledge}
\end{subfigure}
\hfill
\begin{subfigure}{0.35\textwidth}
    \includegraphics[width=\textwidth]{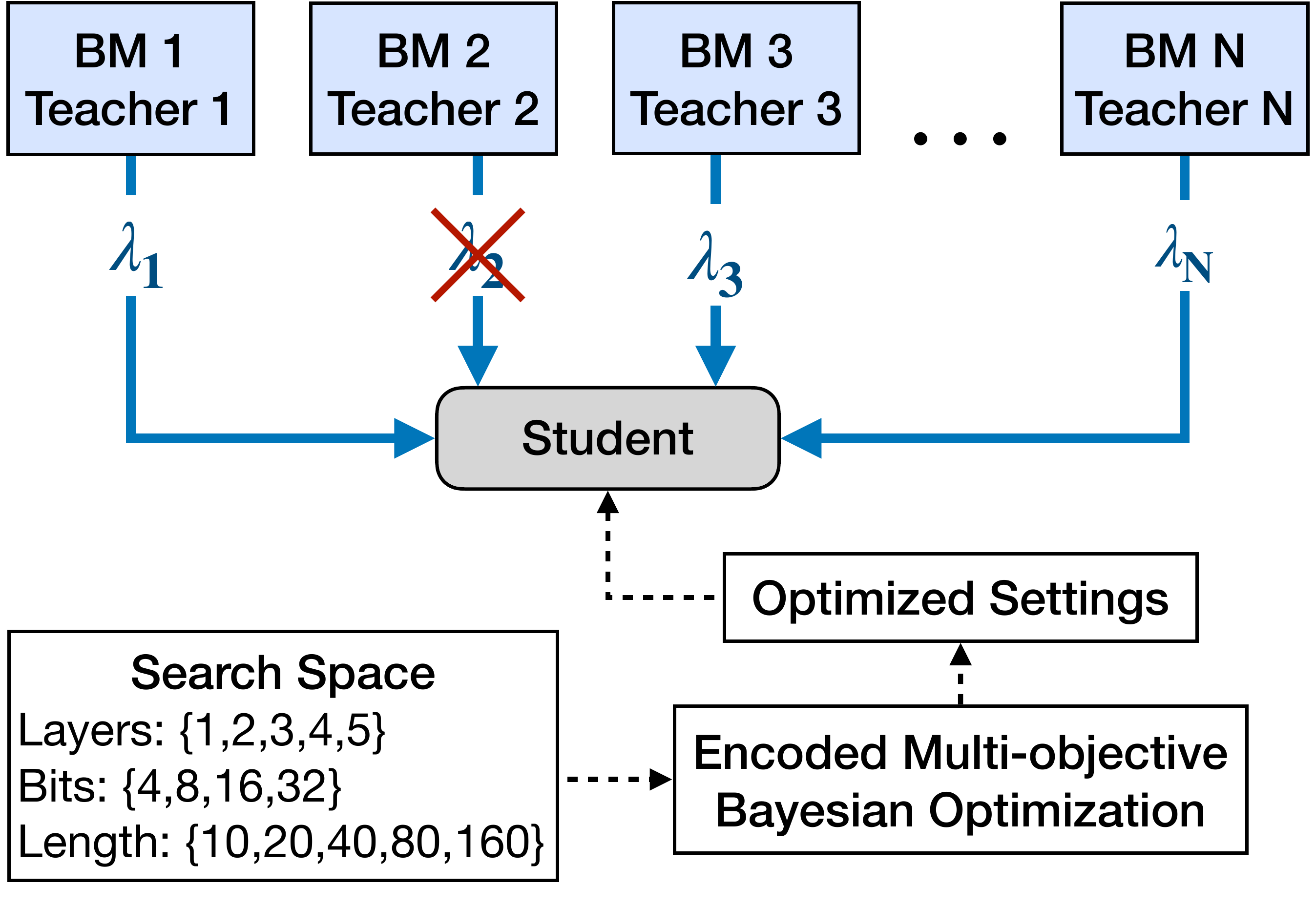}
    \caption{LightTS.}
    \label{subfig:lightts}
\end{subfigure}
\vspace*{-3.09cm}

\vspace*{-0.5em}
\caption{Ensemble Classifier, Knowledge Distillation, and LightTS. 
}
\label{fig:figures}

\end{figure*}

Time series classification methods are already capable of high accuracy~\cite{RuizFLMB21}. 
Most state-of-the-art methods~\cite{MiddlehurstLFLB21,LubbaSKSFJ19,MiddlehurstLB20} rely on \emph{ensemble} learning, where multiple base models perform classification jointly. 
Ensemble learning 
is based on the wisdom of the crowd, suggesting that a joint result from the crowd is often 
superior to any result from a single source~\cite{AggarwalS17}. 
%
As shown in Figure~\ref{subfig:ensemble}, $N$ base models ($BM$s), e.g., classifiers based on InceptionTime~\cite{FawazLFPSWWIMP20}, Temporal Dictionary~\cite{MiddlehurstVB19}, or Forest Classifier~\cite{DengRTM13}, are combined to provide a joint result. 
%
A simple and effective combination approach that is used commonly is to assign identical weights to all base model, e.g., $1/N$ in Figure~\ref{subfig:ensemble}, 
such that all models contribute equally to the result of the ensemble~\cite{Dietterich00,ZhouWT02,LargeLB19,CamposKGHZYJ21}.

While ensemble based methods 
achieve state-of-the-art accuracy, the use of multiple base models requires significant resources 
for storing and executing the base models, 
%
which prevents their use in resource-limited environments, 
%
e.g., on edge devices. For example, the ongoing transformation of power grids to support sustainable energy sources, e.g., wind and solar, relies on power electronics (PE) devices that often have limited memory and computational capabilities. It is of great interest to be able to perform time series classification on PE devices---classifying workload time series into different load levels can enable adaptive control and maintenance~\cite{Zhao21}.
To enable such uses
, it is important to develop lightweight models, e.g., using quantized (e.g., 4, 8, or 16 bits) parameters instead of full-precision (i.e., 32 bits) parameters. For instance, a 16-bit quantized model may use only 50\% of the storage of its full-precision 32-bit counterpart. 
Focusing on the domain of time series classification 
enables demonstrating the applicability of the proposed method in real-world use cases such as PE adaptive control, 
in addition to enabling evaluation of the proposal. Our methodology is applicable in other domains with minor adjustments. 

One approach to building such lightweight models in the setting of large ensembles  
is Knowledge Distillation~\cite{HintonVD15}. 
The idea is to treat the large ensemble as a \emph{Teacher} and then train a lightweight \emph{Student} model to mimic the results from the teacher---cf.\ Figure~\ref{subfig:knowledge}. 
Though being an effective approach, two main challenges remain. 

\noindent
\textbf{Lack of Distillation Flexibility: } An important design consideration in ensemble learning is to ensure a high level of diversity among the base models. 
However, existing knowledge distillation employs a single distillation step that employs the ensemble's combined output to teach the lightweight model, causing the contribution by each base model to be fixed. When an ensemble uses equal weights to combine the results of its base models, all base models contribute equally to the lightweight model. 
This significantly reduces the distillation flexibility. 
For example, some base models may be more suitable for guiding the training of a 4-bit quantized student model, while other base models may be more suitable for a 16-bit model. Existing solutions are unable to identify and exploit such diversity in the distillation process. 
As an analogy, in a real-world teacher-student context, it is a good strategy to align different subject teachers, e.g., physics or math teachers, with the students who lack of the corresponding knowledge, e.g., physics or math.  
Thus, using the same teacher, i.e., the ensemble, to teach different quantized student models is too rigid. To enhance distillation flexibility, an approach is needed that can select different base models adaptively, disregarding some models, and can assign appropriate weights to the selected models, thereby distilling knowledge according to each student model's level of compression.

\begin{figure}[ht!]
\small
    \edef\windows{"","","","U:l=4,b=4","","","V:l=5,b=8","","",""}
    \begin{tikzpicture}
        \begin{axis}[
            xlabel=Model Size,
            ylabel=Accuracy,
            ymin=0,
            ymax=1,
            xmin=50000,
            xmax=190000,
            scaled x ticks=base 10:-3,
            xtick scale label code/.code={},
            xticklabel={\pgfmathprintnumber{\tick} K},
            xtick = {60000,80000,100000,120000,140000,160000,180000},
            xticklabels = {60 K,80 K,\textbf{100 K},120 K,\textbf{140 K},160 K, 180 K},
            width=1*\linewidth,
            height=0.48*\axisdefaultheight,
            legend pos= outer north east]
            \addplot[blue,fill=blue,mark=*, fill opacity=0.2] table[x=Cost, y=Single] {figures/data/pareto_example.txt}\closedcycle;
            \addplot[red,only marks,mark=*,mark size=3pt,nodes near coords=\pgfmathsetmacro{\winstring}{{\windows}[\coordindex]}\winstring, nodes near coords style={text=black,opacity=1,font=\small,xshift=-5pt,yshift=15pt,anchor=north}] table[x=Cost, y=Choice] {figures/data/pareto_example.txt};
            \addplot[black,only marks,mark=x,mark size=3pt] table[x=Cost, y=Under] {figures/data/pareto_no_opt.txt};
            \addplot [black, mark=none, thick=2pt] coordinates {(100000, 0) (100000, 1)};
            \addplot [red, mark=none, thick=2pt] coordinates {(140000, 0) (140000, 1)};
        \end{axis}
          \begin{axis}[
            width=1*\linewidth,
            height=0.48*\axisdefaultheight,
            ymin=0,
            ymax=1,
            xmin=50000,
            xmax=190000,
            scaled x ticks=base 10:-3,
            xtick scale label code/.code={},
            xticklabel={\pgfmathprintnumber{\tick} K},
            xtick = {60000,80000,100000,120000,140000,160000,180000},
            xticklabels = {{},{},Device \#1,{},Device \#2,{}, {}},
            axis x line*=top]
            hide y axis]
          \end{axis}
    \end{tikzpicture}
    \caption{Pareto Frontier Example. 
    Circles and crosses represent possible student settings. The Pareto frontier consists of circles that represent optimal settings, as no other settings exist with higher accuracy and smaller model size.
    }
    \label{fig:pareto_example}
\end{figure}
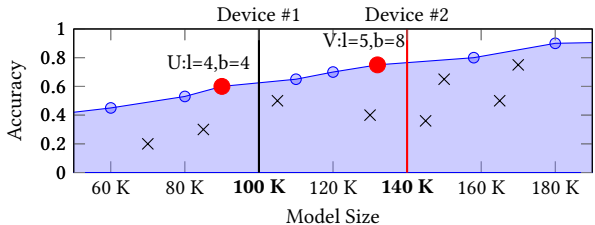

\noindent
\textbf{Lack of Support for Pareto Optimal Settings: }
Existing knowledge distillation considers distillations under specific student settings~\cite{abs-2006-03669}. For instance, for a deep learning based student model, a student setting often specifies the number of layers and the quantization bits of the parameters, e.g., 3 layers, 8 bits, and 40 filter length as shown in Figure~\ref{subfig:knowledge}. 
However, in practical settings, we often do not know how to configure such specific student settings for different devices. Instead, only storage constraints 
are known.
Thus, it is beneficial to obtain Pareto optimal settings considering accuracy vs. model sizes, a.k.a.\ Pareto frontiers (see the example in Figure~\ref{fig:pareto_example}) over a large number of possible student settings. Given edge devices with different storage constraints, Pareto frontiers facilitate the selection of model settings with the highest accuracy while meeting the storage constraints. 
For example, consider Device \#1 with a memory constraint of 100K. The Pareto frontier in Figure~\ref{fig:pareto_example} 
implies that Model $U$ is the optimal model for the device, as it has the highest accuracy among all models whose size is within 100K. The settings of Model $U$, 4 layers and 4-bit parameters, should be used for Device \#1. 
Similarly, for Device \#2 with a memory constraint of 140K, 
we can use the frontier to identify an optimal model, i.e., Model $V$. 
Thus, an effective and efficient method to identify the Pareto frontier from a large number of student settings is called for since the search space can reach sizes of $10^5$ or even more. 

To address the two limitations, 
we propose \texttt{LightTS}, a flexible framework to obtain \underline{Light}weight \underline{T}ime \underline{S}eries classification models using novel adaptive ensemble distillation, as illustrated in Figure~\ref{subfig:lightts}. 
\texttt{LightTS} is flexible in that it is model-agnostic---different types of base models, including both deep and non-deep learning, can be used in an ensemble that serves as input to \texttt{LightTS}. 
%

\noindent
\textbf{Addressing Challenge 1:} 
To ensure high distillation flexibility, we propose a novel \emph{adaptive ensemble distillation}. \texttt{LightTS} treats all base models as teachers, rather than using the ensemble as a single teacher, as done in classic knowledge distillation. 
In addition, \texttt{LightTS} adaptively assigns appropriate weights to different teachers ($\lambda_i$ for teacher $i$ in Figure~\ref{subfig:lightts}). Given a specific quantized student model, e.g., a 16-bit model, \texttt{LightTS} enables assignment of higher weights to the teachers that may contribute more knowledge to the training of the 16-bit model. It is also possible to disregard unimportant teachers by assigning them zero weight. When a differently quantized student model is called for, \texttt{LightTS} is able to assign a different set of weights to the teachers. This offers considerable distillation flexibility. We achieve this by a novel bi-level optimization approach---an inner level optimization learns the quantized model parameters of the student model, while an outer level optimization adjusts the teacher weights $\{\lambda_i\}_{i=1}^{N}$. 

\noindent
\textbf{Addressing Challenge 2:} 
We propose a novel encoded multi-objective Bayesian optimization method to identify the Pareto frontiers (see the lower part in Figure~\ref{subfig:lightts}). 
To do so, we first define a search space that models a wide variety of quantized student model settings. 
As shown in Figure~\ref{subfig:lightts}, the search space specifies possible layers per block (e.g., 1, 2, 3, 4, or 5),  quantized parameter bit-widths (e.g., 4, 8, 16, or 32 bits), and filter length (e.g., 10, 20, 40, 80, or 160). Each layer can use parameters of a distinct size, thus giving rise to a wide variety of quantized model settings. 

Given a quantized model setting, we are able to obtain its accuracy, by using the proposed adaptive ensemble distillation, along with its model size, by counting the total bits. However, it is prohibitively inefficient to compute the accuracy and size of all quantized student model settings in the search space. Instead, we propose a novel encoded multi-objective Bayesian optimization method to explore the most promising settings while considering both accuracy and model size. 
In particular, we propose a novel encoding scheme for the search space such that the encoded space captures appropriate semantics and fits Bayesian optimization better. This enables effective identification of Pareto frontiers.

\noindent
\textbf{Contributions:}
To the best of our knowledge, this is the first study that 
introduces an adaptive ensemble distillation scheme and a generic framework to identify a set of Pareto optimal lightweight models that comply with different storage constraints for time series classification. 
The paper makes the following contributions. 
\begin{itemize}
    \item It proposes an ensemble distillation strategy that is able to adaptively select important base models and assign these appropriate weights, while disregarding unimportant base models, thus enabling more flexible knowledge distillation to lightweight models than hitherto possible.
    \item It proposes a novel encoding scheme along with an encoded multi-objective Bayesian optimization method to find Pareto frontiers, facilitating the identification of optimal models under different storage constraints.
    \item It reports on comprehensive experiments that justify key design decisions and demonstrate that \texttt{LightTS} is able to outperform state-of-the-art solutions. 
\end{itemize}
The remainder of the paper is organized as follows. Section~\ref{sec:preliminaries} covers preliminaries. Section~\ref{sec:method} details the framework, and Section~\ref{sec:experiments} reports on the experiments. Section~\ref{sec:related_work} reviews related work, and Section~\ref{sec:conclusions} concludes.
\section{Preliminaries} \label{sec:preliminaries}

This section presents concepts that are necessary to introduce the proposed framework.
\subsection{Time Series Classification}

\subsubsection{Time Series}
A time series $\mathcal{T} = \langle \mathbf{t}_{1}, \mathbf{t}_{2}, \dots, \mathbf{t}_{C} \rangle$ is a sequence of $C$ observations where each observation is an $M$-dimensional vector, so $\mathbf{t}_{j} \in \mathbb{R}^{M}$. 

\subsubsection{Labeled Time Series}
A labeled time series set $\mathcal{D}$ is a collection of $(\mathcal{T}_i,l_i)$ pairs, where $\mathcal{T}_i$ is a time series and $l_i\in \mathbb{L}$ is the time series's label, indicating a specific class among a set of classes $\mathbb{L}$. %
For example, in a human activity time series data set, $\mathbb{L}$ includes labels representing different activity classes, such as walking, jumping, and jogging. 

\subsubsection{Time Series Classification}
A time series classifier is a function that takes as input a time series and returns its corresponding label. 
During training, we are given a labeled time series set to train the classifier. The accuracy of the classifier is evaluated on time series that do not appear in the labeled time series set using for training. 

\subsection{The \texttt{InceptionTime} Classifier} \label{subsec:inception}
We use \texttt{InceptionTime}~\cite{FawazLFPSWWIMP20}, a state-of-the-art neural classifier for time series, as the student model in \texttt{LightTS}.
%
Two considerations lead to this choice of a neural classifier for achieving lightweight student models.
First, neural classifiers 
achieve state-of-the-art 
time series classification accuracy~\cite{FawazLFPSWWIMP20}. 
%
%
Second, neural classifiers offer a high degree of compressibility to be discussed 
in Section~\ref{sec:quantized}. 

\begin{figure}[ht!]
    \centering
    \includegraphics[width=\linewidth]{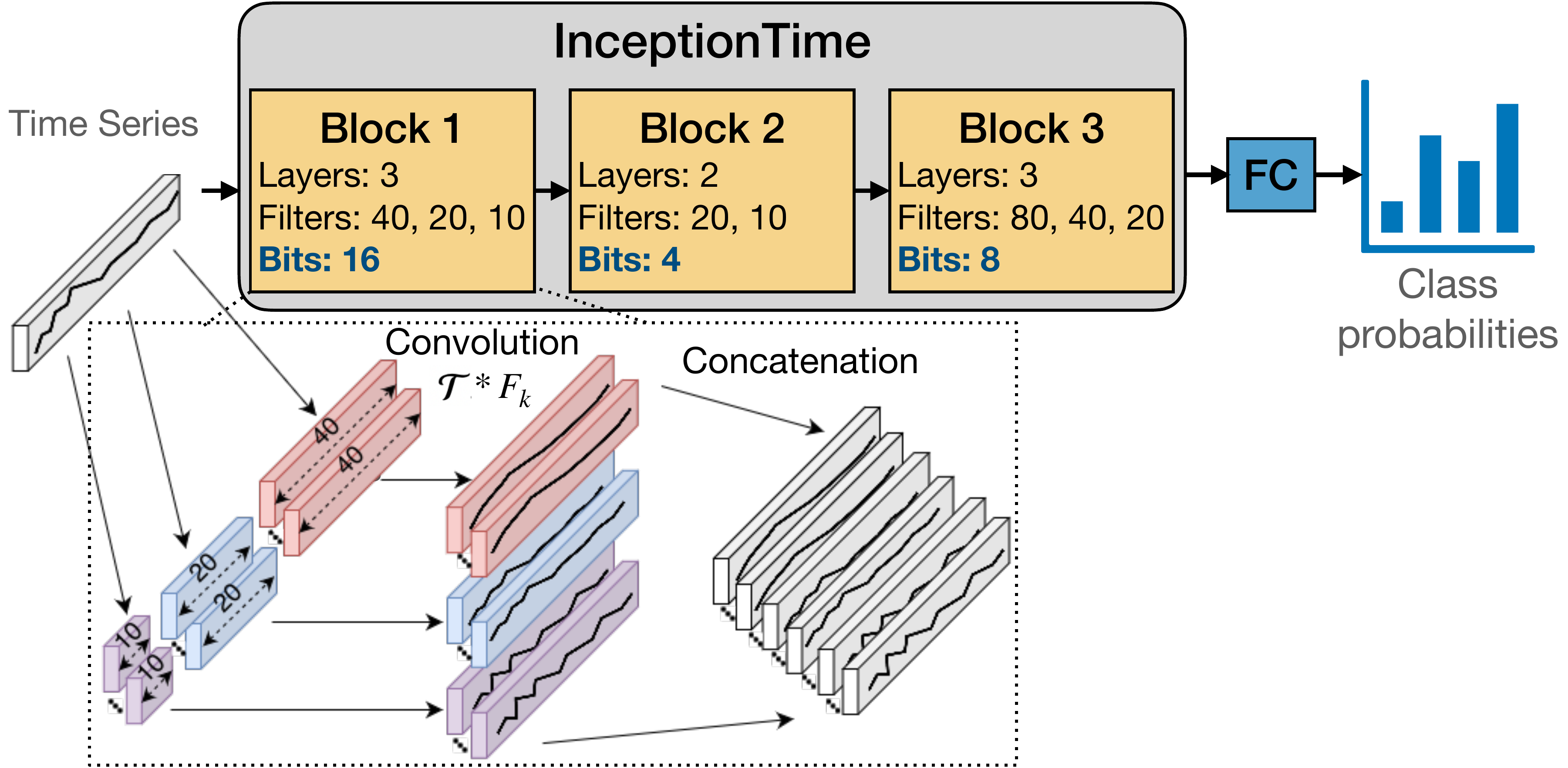}
    \caption{\texttt{InceptionTime} Classifier.}
    \label{fig:inception}
\end{figure}

%
\texttt{InceptionTime} is a convolution neural network that employs variable\Hyphdash length convolution filters 
to capture temporal patterns of different time spans.
Figure~\ref{fig:inception} shows an \texttt{InceptionTime} model with 3 blocks, 
each with multiple convolutional layers  (cf. the zoom-in of the first block). A layer uses a specific number of same-length convolution filters to extract features from the time series $\mathcal{T}$. Then, the different convolution layers are stacked. Specifically, $\mathcal{T}^{(i)}= \concat_k \mathcal{T}^{(i-1)} \ast F_k$ where $F$ is a convolution filter, $\ast$ denotes 1D convolution, $\concat$ denotes concatenation, and $\mathcal{T}^{(i)}$ is the output of the $i$-th block and $\mathcal{T}^{(0)}=\mathcal{T}$.
The lengths of convolution filters in different layers generally vary, enabling the 
capture of patterns of varying lengths. More specifically, the filter lengths are decreased by half. For the 3-layer block, the filter lengths may be 40, 20, and 10. 
Processing time series with multiple layers with different filters provides a comprehensive context for classifying time series since patterns of different length will fit better on the filters with the closer length, as their receptive field~\cite{LuoLUZ16} matches the pattern. Thus, the concatenated output of the layers is the input for the following block.
Finally, the output of the last block is passed to a fully connected (FC) layer with a Softmax function to assign a class distribution to the time series. 

\subsection{Quantized Neural Classifiers} \label{sec:quantized}

\begin{figure}[ht!]
    \centering
    \includegraphics[width=0.7\linewidth]{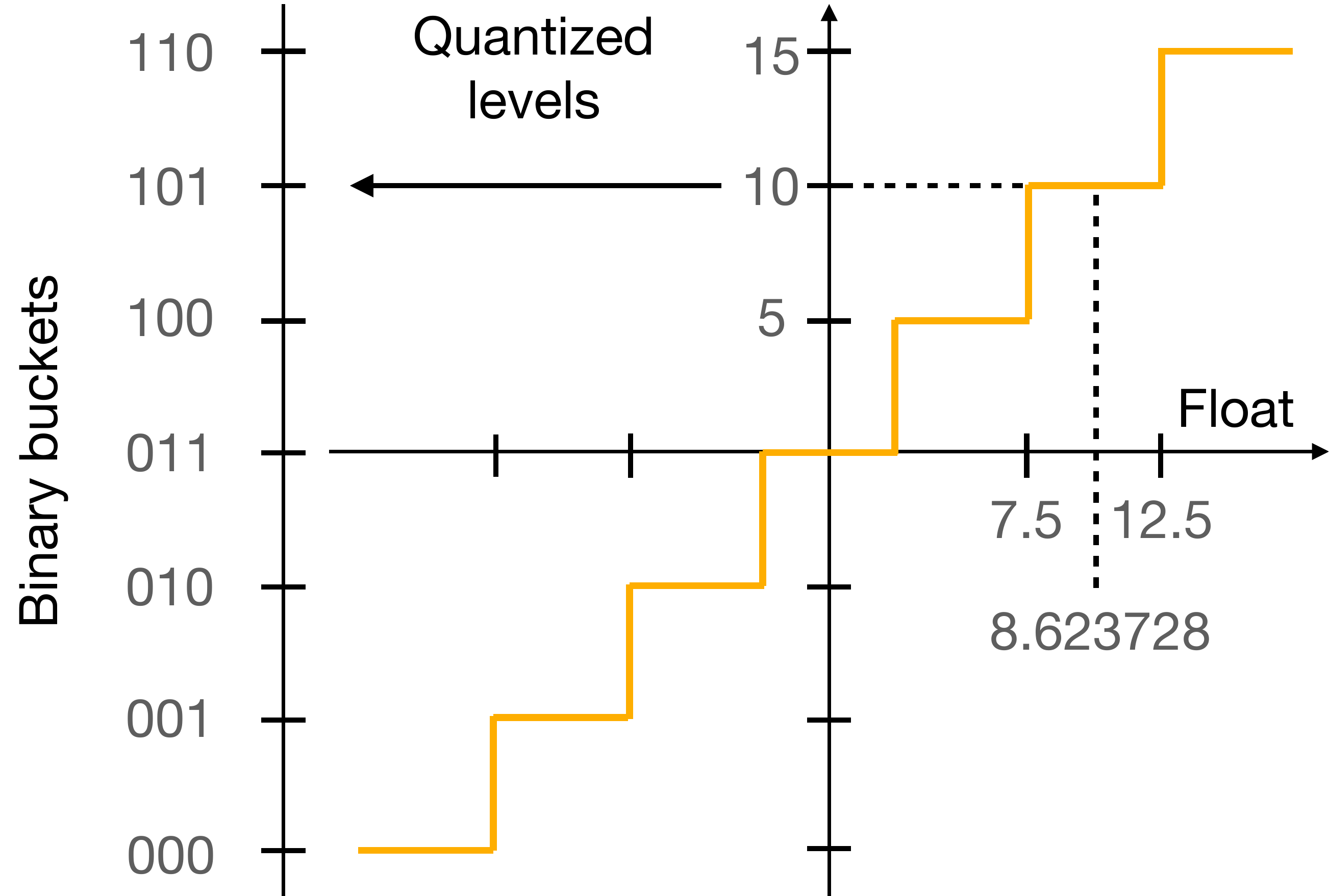}
    \caption{Quantization Mapping. }
    \label{fig:quantization}
\end{figure}

Neural classifiers offer a high degree of compressibility. 
A neural classifier like \texttt{InceptionTime} often consists of blocks, each with multiple layers, each of which in turn includes convolutional filters with different lengths.  
The parameters in the filters are often 32-bit float, so full-precision \texttt{InceptionTime} always use 32-bit parameters.
Thus, to compress a neural classifier, it is possible to reduce the number of blocks, the number of layers in a block, the filter lengths, and the bit-width of the parameters in the different layers (i.e., parameter quantization), as shown in Figure~\ref{fig:inception}. 
{
For example, Figure~\ref{fig:quantization} illustrates how full-precision parameters, i.e., 32-bit floats, can be quantized into 3-bit parameters using uniform quantization~\cite{GongLJLHLYY19}.  
As 8.623728 falls into the interval [7.5, 12.5), it maps to 10, which is then mapped to the 3-bit 
bucket 101. 
}

\subsection{Knowledge Distillation} \label{sec:knowledge_other}
Knowledge Distillation (KD)~\cite{HintonVD15} aims to transfer knowledge from a teacher model to a student model, where the teacher is often a larger model with higher discriminative capacity than the student. 
In classification, the knowledge is represented by a probability distribution over classes produced by the teacher model. 
Let $q$ and $p$ represent the class distributions from the teacher and the student, respectively. Then, knowledge distillation is formalized in Equation~\ref{eq:adjusted_loss}, where $\alpha \in [0,1]$ is a hyper-parameter~\cite{YuanSPLGFJ21}. 
\begin{gather}
    \mathcal{L} = \alpha \times \mathcal{L}_{CE}(p,y) + (1-\alpha) \times {\mathit Dist(q, p)}
    \label{eq:adjusted_loss}
\end{gather}

Specifically, the loss function is computed over two components that are weighted by $\alpha$.
The first component is the cross-entropy (CE) between the student class probabilities $p$ and the ground truth label $y$, which provides supervision from the ground truth labels. 
The second component represents the distance between the teacher and student distributions $q$ and $p$, e.g.,  Kullback–Leibler (KL) divergence, to encourage a student to mimic the behavior of a more powerful teacher.
They both contribute to training an accurate student.

\begin{figure}[ht!]
    \centering
    \includegraphics[width=0.78\linewidth]{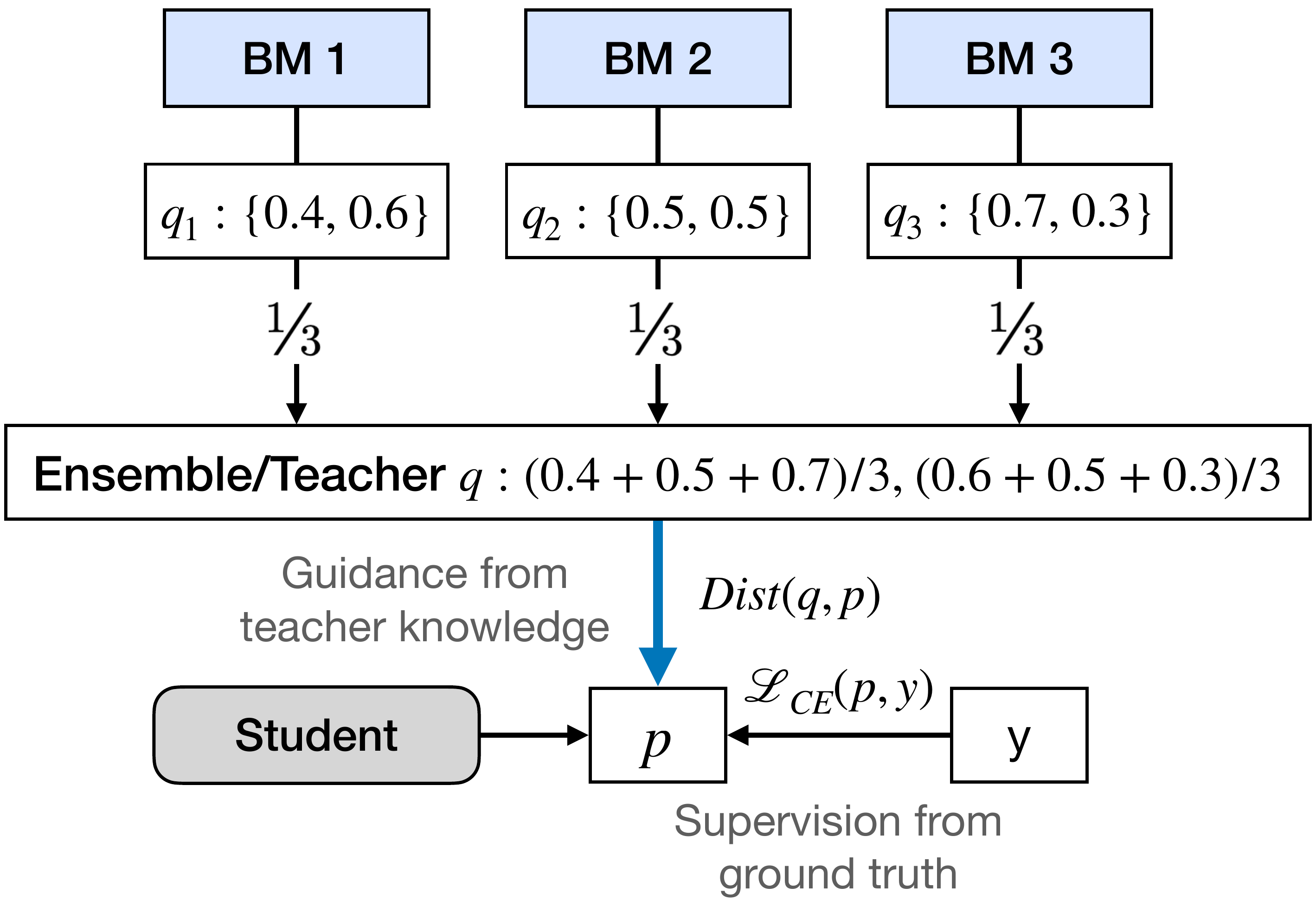}
    \caption{Classic Ensemble Knowledge Distillation.}
    \label{fig:ensemble_basic}
\end{figure}

When a group of base models is available as the teachers, their average-ensemble, $q=1/N \times \sum q_i$, 
where $q_i$ is the class distribution returned by the $i$-th base model, is typically considered as the knowledge source \cite{DuYLW00Z20}.
Figure~\ref{fig:ensemble_basic} shows an example of the knowledge distillation with an ensemble consisting of three base models. 

\section{L\texorpdfstring{\lowercase{ight}}-TS} \label{sec:method}


We first introduce the two problem scenarios supported by \texttt{LightTS}. Then, we proceed to introduce two main building blocks of \texttt{LightTS}, \emph{adaptive ensemble distillation} and \emph{Pareto frontier identification}, which each targets one of the two problem scenarios. 

\subsection{Problem Scenarios} \label{sec:scenarios}

\begin{figure}[ht!]
    \centering
    \includegraphics[width=0.87\linewidth]{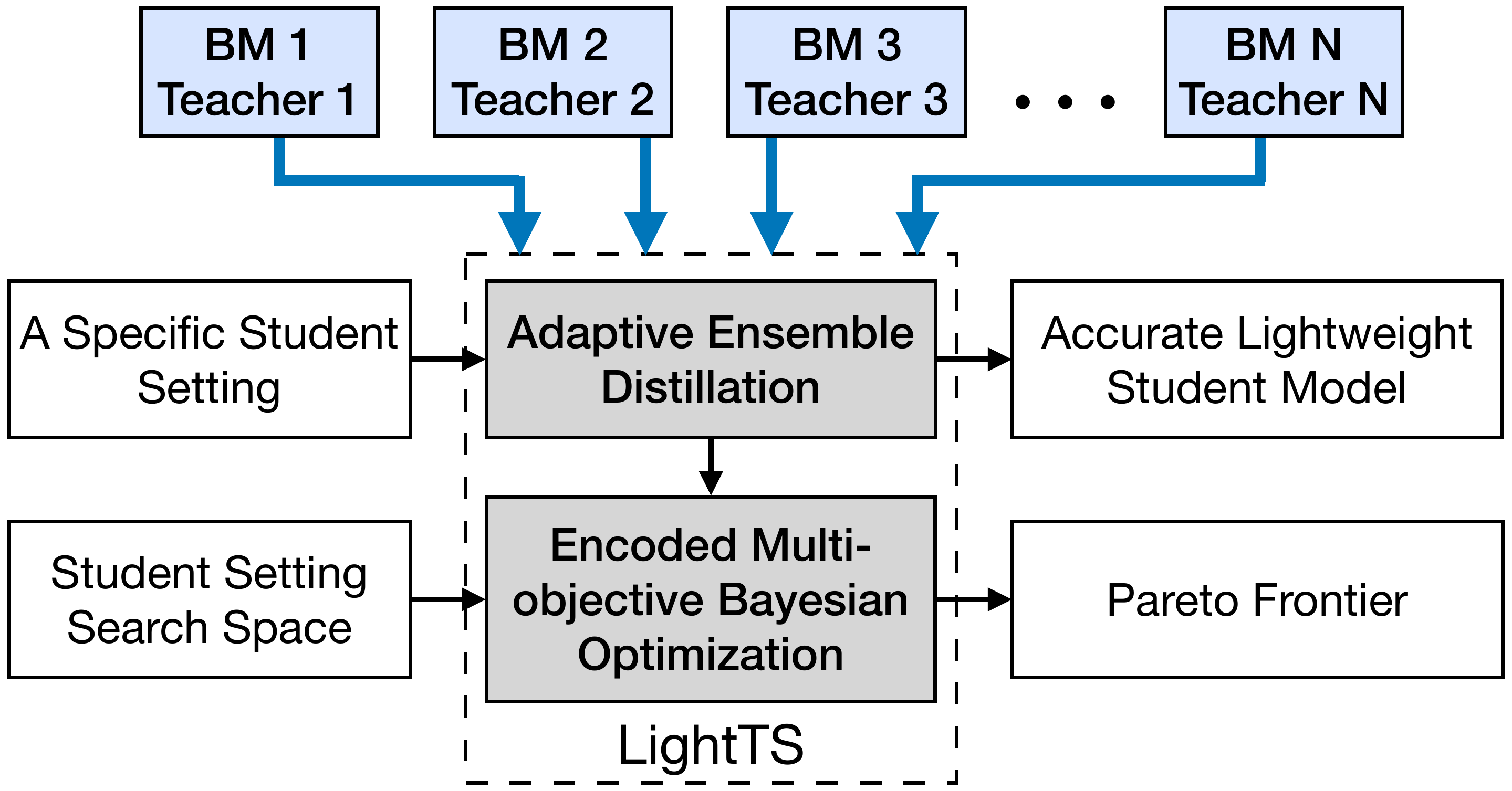}
    \caption{\texttt{LightTS} Workflow.} 
    \label{fig:workflow}
\end{figure}

The input to \texttt{LightTS} is an already trained ensemble consisting of $N$ full-precision base models $\{\mathit{BM}_i\}_{i=1}^{N}$, as shown in Figure~\ref{fig:workflow}. 
Although we employ \texttt{InceptionTime} as the quantized student model, \texttt{LightTS} is not limited to 
%
\texttt{InceptionTime} but can also support other bases models, including both deep and non-deep learning models. 
It is only required that the base models output class distributions. 
This design makes \texttt{LightTS} a generic framework.

The two problem scenarios supported by \texttt{LightTS} are 
shown in Figure~\ref{fig:workflow}.
First, 
a specific lightweight student setting is given, e.g., the number of layers, filters length, and the quantization bits per layer. Here, the goal of \texttt{LightTS} is to build an accurate student model under this lightweight setting. 
We propose \emph{adaptive ensemble distillation} to support this scenario (cf. Section~\ref{sec:adaptive}). 

However, establishing student settings 
that achieve good accuracy vs. space trade-offs is non-trivial, and  
it is inefficient to manually identify such promising settings. 
%
Therefore, we consider a second problem scenario, where
the setting for the quantized student model is not given. Instead, we define a search space that covers a wide variety of lightweight student settings. Here, the goal of \texttt{LightTS}
is to identify 
%
the Pareto frontier in this space, which includes Pareto optimal settings with the property that no other settings have higher accuracy and smaller model size. 
We propose an encoded multi-objective Bayes optimization method to identify the Pareto frontier (cf. Section~\ref{sec:pareto}). 
Then, given a device with a specific memory constraint, we can choose the setting that achieves the highest accuracy while complying with the memory constraint. 

\subsection{Adaptive Ensemble Distillation} \label{sec:adaptive}

In the first problem scenario where 
a lightweight student setting is given, we propose an \emph{adaptive ensemble distillation} (\texttt{AED}) process which assigns the teachers with learnable, adaptive weights, such that different teachers can contribute differently to the student. 
%
The scenario is aligned with the real-world analogy of taking advantage from the skills of particular teachers to prepare the students who require them the most, such as focusing on a specific subject teacher. 

\subsubsection{Learning adaptive weights} \label{sssec:learningAED}
\begin{figure}[ht!]
    \centering
    \includegraphics[width=0.78\linewidth]{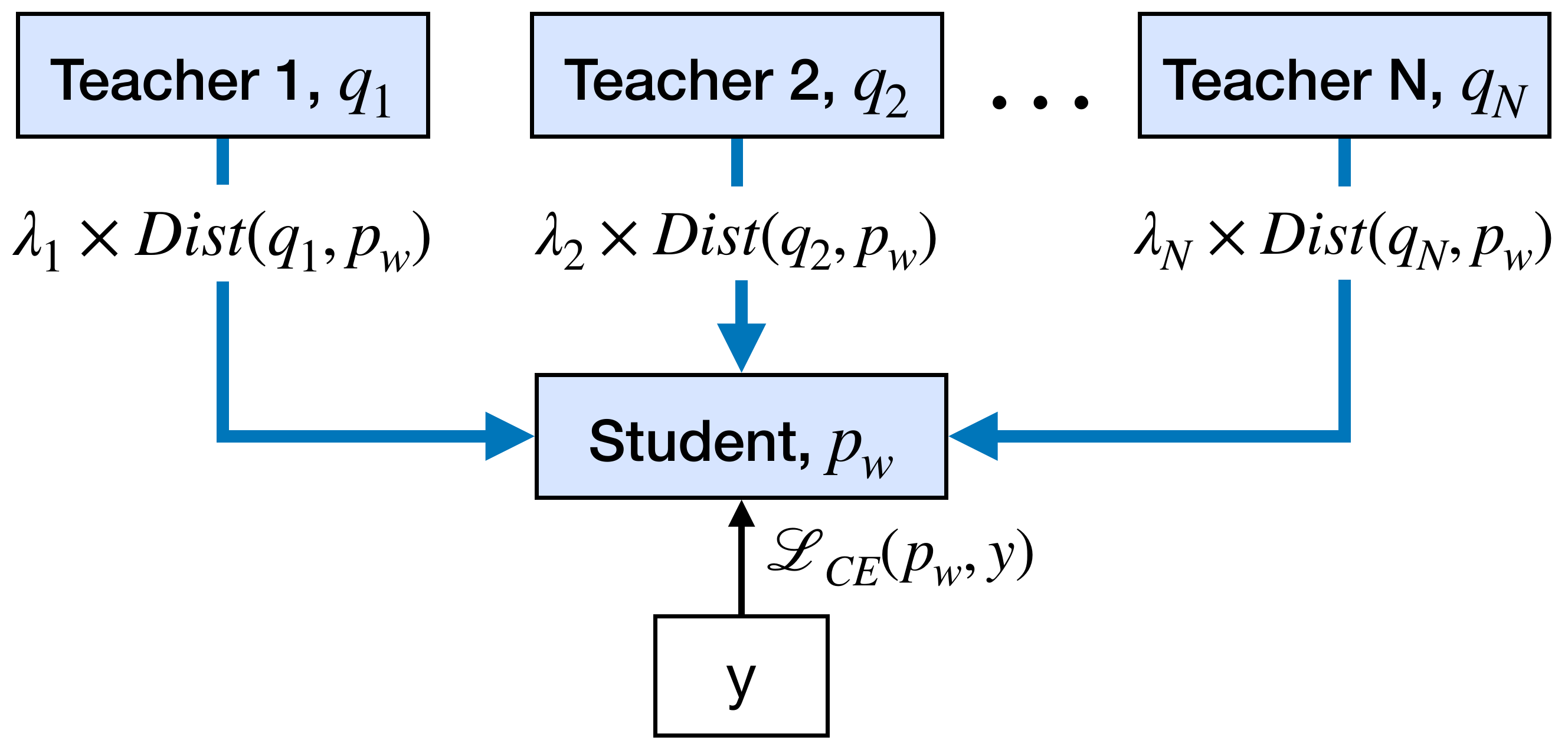}
    \caption{Adaptive Ensemble Distillation (\texttt{AED}).}
    \label{fig:knowledgeDistillation}
\end{figure}

%
To obtain an accurate lightweight student, we need to consider two perspectives. First, we need the supervision from the ground truth labels in the training data. Second, we need the guidance from the knowledge of the already trained full-precision teachers. 
To distill knowledge from an ensemble of already trained full-precision teachers, the classic knowledge distillation (cf. Figure~\ref{fig:ensemble_basic} in Section~\ref{sec:knowledge_other}) utilizes 
the average knowledge of the $N$ base models.
%
Instead, in \texttt{AED}, we 
propose to introduce direct connections from every base model to the student, such that each base model works as a teacher, as shown in Figure~\ref{fig:knowledgeDistillation}.
%
%
In the upper part, the knowledge distillation is drawn by a direct link between every teacher and the student. The knowledge from each teacher, i.e., its class distribution $q_i$, guides the learning of the student's class distribution $p_w$, through a distance metric ${\mathit Dist(q_i, p_w)}$, 
controlled by a weight $\lambda_i$.
In the bottom, the 
student's class distribution $p_w$ is also adjusted
with respect to the ground truth $y$ using the cross-entropy ($CE$) loss.
Here, we denote the student's class distribution by $p_w$, meaning that it is the distribution returned by the student with model parameter $w$. 


Formally, in contrast to the ensemble distillation loss shown in Equation~\ref{eq:adjusted_loss}, now the distance among every teacher and the student is included as the second component weighted by their corresponding $\lambda_i \in [0,1]$, where $\sum \lambda_i = 1$, as shown in Equation~\ref{eq:loss_kl}.
Thus, instead of having one consolidated guidance between the ensemble and the student, we now consider individual guidances between the teachers and the student weighted by their associated weight $\lambda_i$. 
\begin{align}
    \mathcal{L} = \alpha \times \mathcal{L}_{CE}(p_w,y) + (1-\alpha) \times \sum^{N}_{i=1}{\lambda_i \times Dist(q_i,p_w)}
    \label{eq:loss_kl}
\end{align}
As it can be observed in the new loss function, the two components represent different types of weights that we want to optimize.
First, the classification cross-entropy loss $\mathcal{L}_{CE}(p_w,y)$ involves $p_w$, which is associated with quantized model parameters $w$ from the student, e.g., quantized convolutional filters.
Second, the degree of contribution $\lambda_i$ comes from each teacher.
Both components are dependent to each other because (i) when $w$ is learned, it requires to know how much the teachers are contributing via their $\lambda_i$; and (ii) while adjusting the set $\boldsymbol\lambda = {\{\lambda_i}\}_{i=1}^{N}$, it requires to perceive the student accuracy, i.e., depending on $w$. 
The condition of having two optimization objectives leads us to consider a bi-level optimization modeling, where each objective can be adjusted alternatively while considering the results from each other.

In the inner-level optimization stage, we learn the model parameters $w$ for the student. While adjusting $w$, the $\boldsymbol\lambda$ is kept static, the distances with respect to the teachers ${\mathit Dist(q_i, p_w)}$ and the classification loss $\mathcal{L}_{CE}(p_w,y)$ can change, as they are dependent on $w$. 
This step uses the training data set.
In the outer-level optimization stage, the $\boldsymbol\lambda$ is adjusted using the validation set, while the distances w.r.t. the teachers and the cross-entropy loss remain unchangeable.
Formally, the above bi-level optimization is detailed next in Equations~\eqref{eq:outer} and \eqref{eq:inner}, where $D_{{\mathit train}}$ and $D_{\mathit validation}$ denote the data for training and validation, respectively. It is detailed in Algorithm~\ref{alg:ensemble_selection}.
\begin{gather}
    \arg \min_{\boldsymbol\lambda}  \mathcal{L}(\boldsymbol\lambda, w^*, D_{\mathit validation}) \label{eq:outer}\\
    s.t., w^* = \arg \min_{w} \mathcal{L}(\boldsymbol\lambda, w, D_{\mathit train}) \label{eq:inner}
\end{gather}
%
Equation~\ref{eq:inner} represents the inner-level optimization.
Under a specific disposition of $\boldsymbol\lambda$, it finds the optimized student weights $w^*$,
meaning that under this specific $\boldsymbol\lambda$, the student with $w^*$ gives the best accuracy on the training data. 
It uses the training set, since in this step it is where the model is supervised trained in terms of classification performance, where cross-entropy loss with ground truths is used.
The step is shown in line~\ref{alg:train_start} in Algorithm~\ref{alg:ensemble_selection}.
Here, $p_w$ represents the class distribution for the student, given its parameters $w$. 
We denote $\lambda_i$ with a box to indicate that it is not changing during the inner optimization step, usually referred as parameters freezing~\cite{BlakeKIW21}. 
Here, the Softmax function $\sigma(\cdot)$ assures $\sum \sigma(\lambda_i) = 1$.

In Equation~\ref{eq:outer}, representing the outer-level optimization, the goal is to find the best  $\boldsymbol\lambda$ given that the student has already an optimal weight $w^*$, so the distances with respect to all teachers are set and weighted by $\boldsymbol\lambda$.
The step uses a validation set to assure that its optimization is independent with respect to the $w^*$ adjustments, and the $\boldsymbol\lambda$ is optimized using back\Hyphdash propagation (BP). 
The $\boldsymbol\lambda$ set acts as hyper\Hyphdash parameters to the inner step, so also using the same training set could derive on unsought conditions such as overfitting, and thus we use the validation set instead. 
The step is shown in line \ref{alg:val_starts} in Algorithm~\ref{alg:ensemble_selection}.
Similarly to line~\ref{alg:train_start}, we use boxes to indicate that the classification cross entropy loss and the distances w.r.t. the teachers are fixed during the outer optimization. 
We run multiple inner-level steps for each outer-level one to have a stable training, meaning that the model will not get different $\boldsymbol\lambda$ values at every iteration.
It gets a single set of $\boldsymbol\lambda$, train for $v$ iterations, and then get another updated set for $\boldsymbol\lambda$.
%

\noindent
\textit{Complexity: }
Algorithm~\ref{alg:ensemble_selection} executes $E$ training epochs. 
In each, BP updates the model parameters $w$; and every $v$ epochs, 
BP updates 
$\boldsymbol\lambda$. 
This gives cost 
$E \times BP_w + E/v \times BP_{\boldsymbol\lambda}$, where $BP_w$ and $BP_{\boldsymbol\lambda}$ are the costs for updating the parameters $w$ and $\boldsymbol\lambda$ using BP, respectively. 
Since there are many more parameters in $w$ than in
$\boldsymbol\lambda$ and $v$ is a small constant, we obtain an asymptotic complexity 
of $\mathcal{O}(E \times BP_w)$, the same as for 
%
classic Knowledge Distillation. 

\algblock{Input}{EndInput}
\algnotext{EndInput}
\algblock{Output}{EndOutput}
\algnotext{EndOutput}
\begin{algorithm}
\caption{Learning Adaptive Weights}
\label{alg:ensemble_selection}
\begin{algorithmic}[1]

\Input : $[q_1, q_2, \dots, q_N]$: $N$ class distributions from full precision teachers.
\EndInput
\State $[\lambda_1, \lambda_2, \dots, \lambda_N]\gets \text{ Uniform weight initialization: } 1/N$
\State $v \gets \text{ Validation steps}$
\State Quantized weight $w$: $\gets$ Random initialization 
    \For {training epochs $e \gets 1,2,\ldots, E$}
        \\ \Comment{Using training data}
        \State $\begin{array}{l}\arg \min_{w} \alpha \times \mathcal{L}_{CE}(p_w, y) \, +  \\ 
        \qquad \qquad (1-\alpha) \times \sum^{N}_{i=1}{\boxed{\sigma(\lambda_i)} \times {\mathit Dist(q_i,p_w)}}\end{array}$ \label{alg:train_start}
        \\
        \If {$e \mod v = 0$}  \Comment{Using validation data}
        \State $\begin{array}{l}\arg \min_{\lambda} \alpha \times \boxed{\mathcal{L}_{CE}(p_w, y)}  \ +  \\ 
        \qquad \quad \ \ (1-\alpha) \times \sum^{N}_{i=1}{\sigma(\lambda_i) \times \boxed{{\mathit Dist(q_i,p_w)}}}\end{array}$ \label{alg:val_starts}
        \EndIf
    \EndFor
\end{algorithmic}
\end{algorithm}

In addition to classification, the proposal can  be applied to forecasting by replacing the cross entropy term in Equation~\ref{eq:loss_kl} by a forecasting error term, e.g., mean square error.

\subsubsection{Removing Unimportant Teachers} \label{sssec:remove}
We have teachers and students with different strengths and limitations.
Recall that a design principle of ensemble learning is to maintain high diversity among the base models, i.e., the teachers in our setting. 
The students are also diverse as they are constrained in different ways, e.g., different layers, filter lengths, and bit-widths.
Ideally, we want to associate the students with the teachers that offer them the most knowledge (already achieved in the proposed \texttt{AED}), while removing teachers who are not contributing or even affecting adversely the student learning.
In addition, removing teachers that do not contribute to the student makes it easier to assign more accurate adaptive weights in future iterations.

\begin{figure}[ht!]
    \centering
    \includegraphics[width=0.92\linewidth]{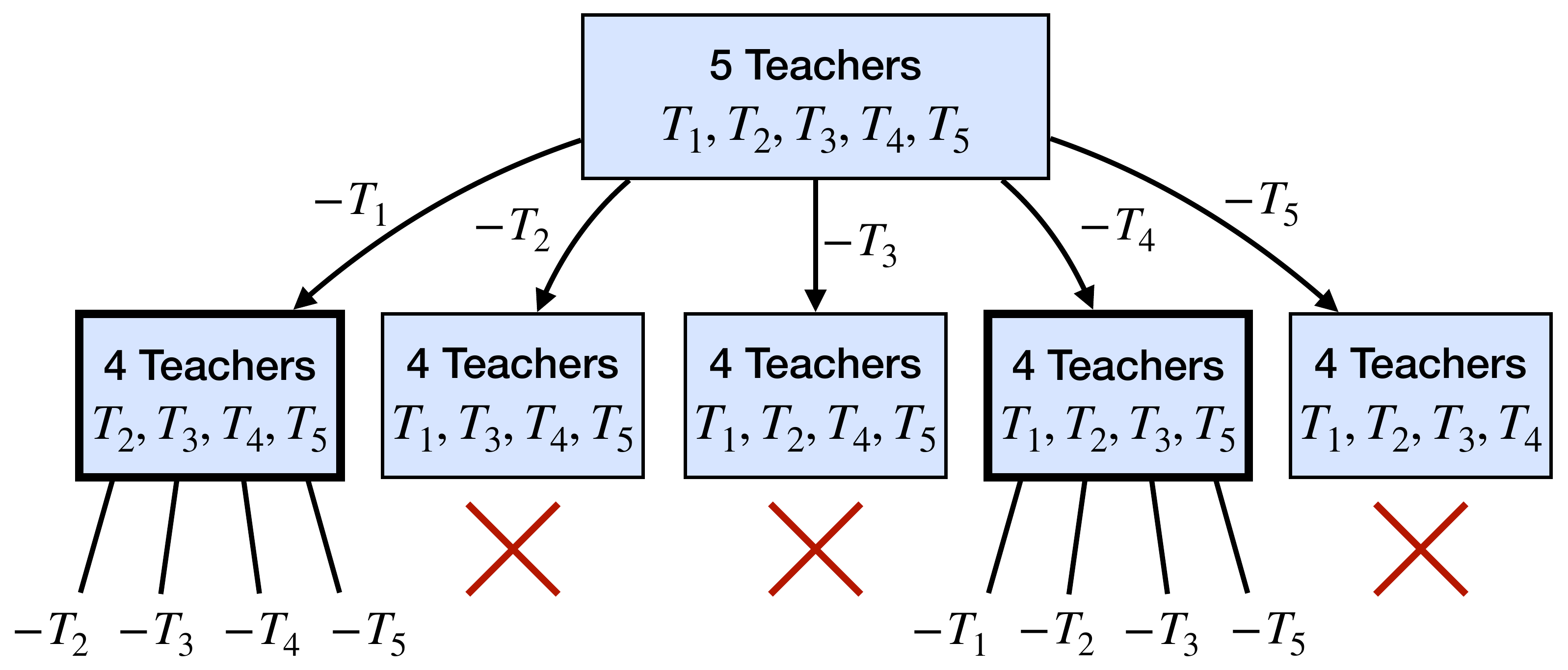}
    \caption{Leave-one-out Based Teacher Removal, Yielding Factorial Growth in the Worst Case.}
    \label{fig:leaveone}
\end{figure}

A simple yet effective principle for removing unnecessary teachers is considering the student classification accuracy as the indicator of improvement. 
%
%
In the current scenario, we can use the student accuracy as the metric for evaluating if the removal of one teacher is contributing to a better guidance to the student.
The criteria can be applied with the leave-one-out idea~\cite{MukherjeeNPR06} that iteratively removes teachers.
%
After removing a teacher, we can check whether the student improves the classification accuracy. 
%
%
%
For example, in Figure~\ref{fig:leaveone} we present a scenario with five teachers. 
If we want to remove one teacher, there are five different ways to do so, i.e., removing $T_1$, $T_2$, $T_3$, $T_4$, or $T_5$.
Then, we can evaluate 
whether removing $T_i$ results in student accuracy improvement. If so, we consider further removing. Otherwise, we stop removing. For example, if removing $T_1$ or $T_4$ further improves the student performance, while $T_2$, $T_3$, and $T_5$ decrease it, we continue the leave-one-out removal for $T_1$ and $T_4$ branches, but stop further removal for the other $3$ cases, as Figure~\ref{fig:leaveone} shows. 

This solution is not efficient if we only rely on the student performance, because we only rely on results but not criteria to choose the teachers that will be removed. 
Therefore, conducting an exhaustive exploration becomes very costly, as shown in Figure~\ref{fig:leaveone}, where leave-one-out of five teachers, marked as $-T_i$, at the first two iterations are exemplified.
%
In the worst case, the number of cases to evaluate following this strategy grows at a factorial pace, so it becomes an intractable problem with relatively few cases, e.g., ten teachers require $10! \approx 3.6 \times 10^6 $ evaluations.

\begin{figure}[ht!]
    \centering
    \includegraphics[width=0.95\linewidth]{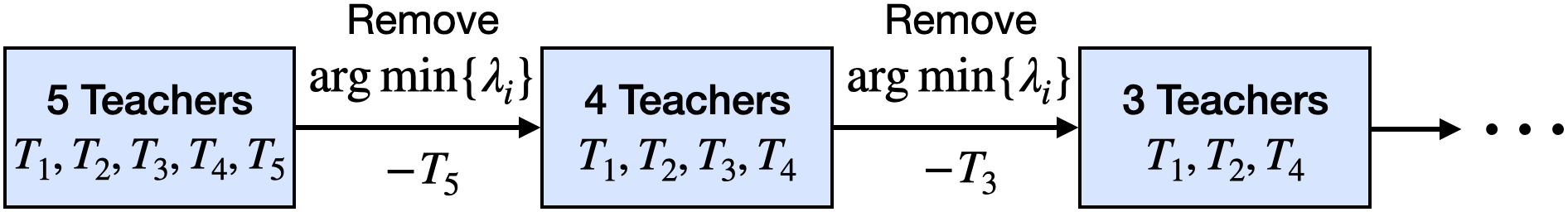}
    \caption{\texttt{LightTS} Teacher Removal.}
    \label{fig:removeteachers}
\end{figure}

%
We propose to utilize the weight $\boldsymbol\lambda$ associated with the teachers to facilitate a more efficient teacher removal strategy. 
%
%
More specifically, the teacher with the lowest $\lambda_i$ is removed.
After removing the teacher, we conduct a new round of \texttt{AED}. 
We keep removing the teacher with the lowest $\lambda$ until running out of base models. 
%
The maximum possible number of iterations is linear, i.e., removing all $N$ teachers. 
Finally, we return the teacher configuration with the highest accuracy. Figure~\ref{fig:removeteachers} shows an example where we remove $T_5$ and then $T_3$.  
%

\begin{figure}[ht!]
    \centering
\begin{subfigure}{0.3\linewidth}
    \includegraphics[width=\textwidth]{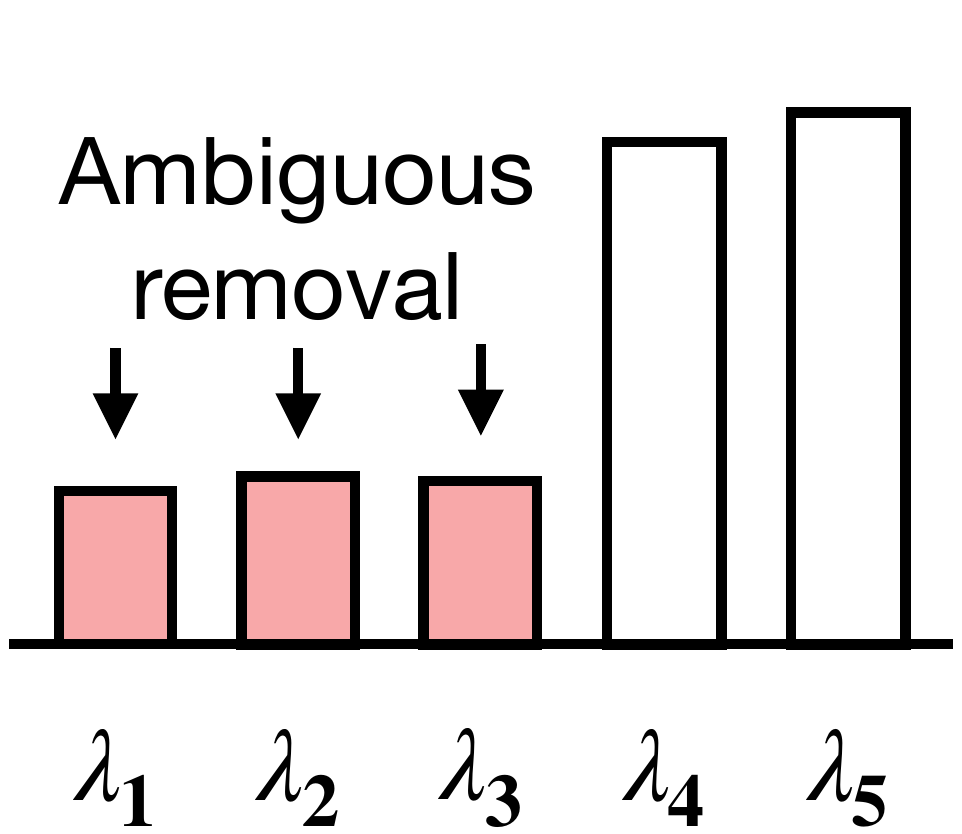}
    \caption{Softmax, Teacher {Importance}.}
    \label{subfig:gumbel1}
\end{subfigure}
\hfill
\begin{subfigure}{0.3\linewidth}
    \includegraphics[width=\textwidth]{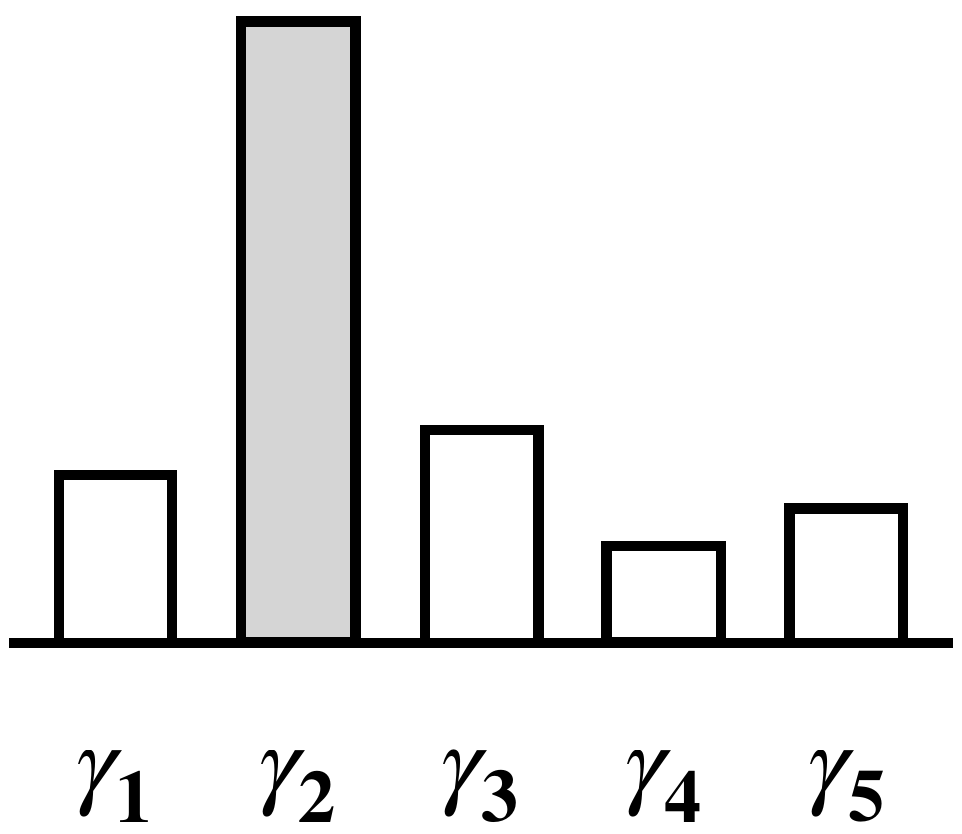}
    \caption{Gumbel, Teacher {Unimportance}.}
    \label{subfig:gumbel2}
\end{subfigure}
\hfill
\begin{subfigure}{0.3\linewidth}
    \includegraphics[width=\textwidth]{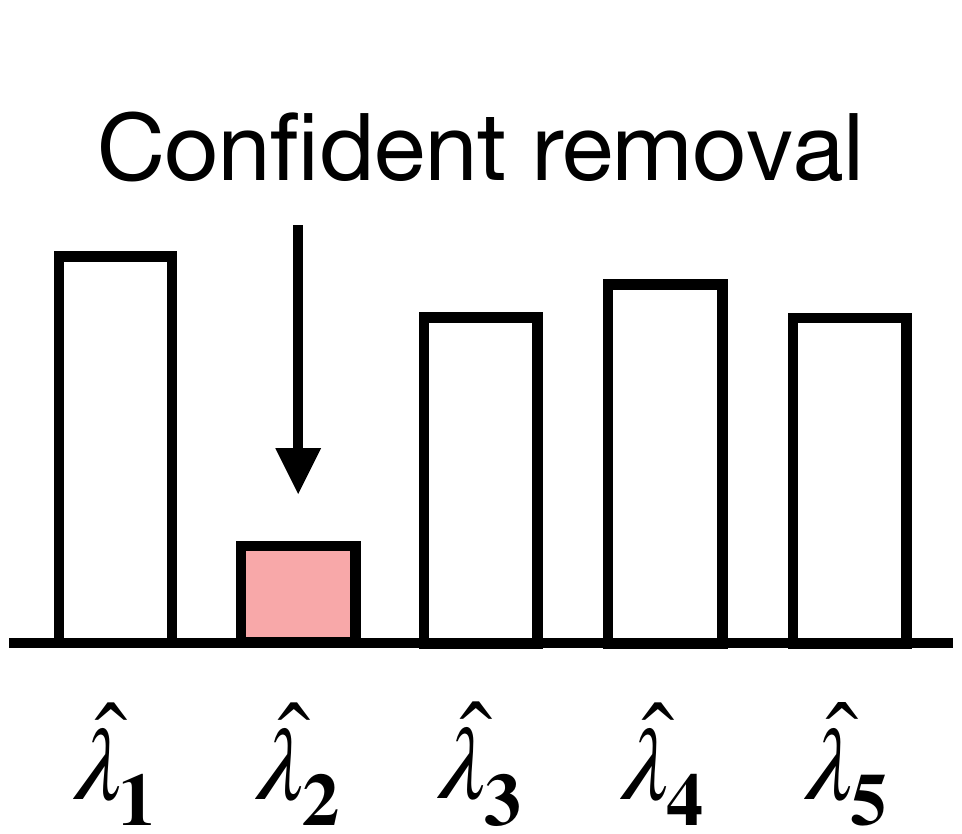}
    \caption{Gumbel, Teacher {Importance}.}
    \label{subfig:gumbel3}
\end{subfigure}
\caption{Confident Teacher Removal.
}
\label{fig:gumble}
\end{figure}

A condition that arises over using $\boldsymbol\lambda$ as the indicator for removing teachers is that, 
after assuring $\sum \sigma(\lambda_i) = 1$ using a Softmax function,
sometimes the $\boldsymbol\lambda$ values become very close.
Thus, it becomes ambiguous which is the weakest teacher to be removed since more than one can be suitable. 
An example of five teachers is shown in Figure~\ref{subfig:gumbel1}, where 
the first three teachers have similarly small $\lambda$ values. 
%

To contend with the above undesired condition, we need a distribution where the majority of weights have similarly large values and only one is significantly smaller, as Figure~\ref{subfig:gumbel3} shows. In such a scenario, the removal decision is more confident, as a clearly weak teacher is removed.

To this end, we consider a reparameterization trick, specifically the Gumbel-Max Trick  \cite{MaddisonTM14,JangGP17}, to introduce stochastic noises $gs$ into weights $\boldsymbol \lambda$ during the teacher removal to make the ``removing'' part differentiable, and leverage a temperature factor $\tau$ to control the sharpness of the reparameterized distribution, $\lambda^*_{i}=\frac{exp((\lambda_i+gs_i)/\tau)}{\sum_j exp((\lambda_j+gs_j)/\tau)}$. In this way, the reparameterized weight contains a deterministic part, i.e., $\boldsymbol \lambda$, and a stochastic part, i.e., $gs$, enabling to sample a teacher to be removed instead of just taking the \textit{argmax}, given their different probabilities. The Gumbel-Max Trick considers the Gumbel distribution~\cite{Gumbel} for noises $gs$ which is stable under operations that involve finding maximums \cite{MaddisonMT17}.
More importantly, through controlling $\tau$, the reparameterized weights smoothly approach the discrete \textit{argmax} computation, so as enlarging the gap between largest weight and the remaining ones and making it likely to become the maximum, as shown in Figure~\ref{subfig:gumbel2}.

Although a Gumbel-Softmax function on $\lambda$ enables the ``removing'' differentiable, it behaves similarly to the \textit{argmax} as the temperature $\tau$ decreases close to zero, while is contrary to our intuition which is supposed to remove the minimal one, i.e., \textit{argmin}. Thus, instead of reparameterizing the ``importance'' of each teacher, we apply the Gumbel-Softmax function to the negative of the set of $\boldsymbol\lambda$ to get the opposite behavior, which we call the ``unimportance'' of teachers 
$\gamma_{i}=\frac{exp((-\lambda_i+gs_i)/\tau)}{\sum_j exp((-\lambda_j+gs_j)/\tau)}$, which is 
 %
shown in Figure~\ref{subfig:gumbel2}.
%
%
Thus, we manage to maximize the teachers unimportance, identifying the teacher that it is most likely to be disregarded based on $\boldsymbol\gamma$, 
as shown in Figure~\ref{subfig:gumbel2}. 
Then, the values are re-parameterized to teacher importance 
by applying a Softmax function to the negative of the unimportance $\hat{\lambda_{i}}=\sigma(-\gamma_{i})$. In $\hat{\lambda}$, the minimal value is confidently identifiable, as Figure~\ref{subfig:gumbel3} shows. 
In Algorithm~\ref{alg:ensemble_selection}, $\hat{\lambda}$ is utilized in lines \ref{alg:train_start} and 
\ref{alg:val_starts}. After Algorithm~\ref{alg:ensemble_selection} finishes, the teacher with the minimal weight $\hat{\lambda}$ is removed. 

\noindent
\textit{Complexity: }
Given an ensemble with $N$ base models, i.e., teachers, we can remove at most $N-1$ teachers. 
This implies that Algorithm~\ref{alg:ensemble_selection} executes at most $N-1$ times. 
Thus, the complexity of \texttt{AED} with teacher removal is 
$\mathcal{O}(N\times E \times BP_w)$. 

\subsection{Identifying Pareto Frontiers} \label{sec:pareto}

In the second problem scenario, instead of identifying a well\Hyphdash performing student for a specific setting, we explore a search space of settings to find optimal sets of student settings that target 
different space restrictions.
To achieve this goal, we first define a search space that fits neural classifiers as well define the notion of the Pareto frontier. Then, we propose an effective encoding scheme, that enables use of a novel multi-objective Bayesian optimization method for identifying Pareto optimal solutions in the search space.

\subsubsection{Search Space} \label{sss:search_space}
The neural classifiers introduced in Section~\ref{sec:quantized} have four components that contribute to the model size of a student: the number of blocks, the number of layers per block, the filter length, and the bit-width. 

The number of blocks controls the network depth. When keeping the number of blocks fixed, changing the number of layers per block also controls the network depth. Thus, we consider a search space that includes only three dimensions: the number of layers per block, the filter length, and the bit-with per block, while keeping the number of blocks $B$ fixed. Different filter length enables the capture of time series patterns of different time spans~\cite{FawazLFPSWWIMP20}, which often vary across data sets. Thus, in addition to contributing to the model size, dynamic adjustment of the filter length enables better classification accuracy.

Given $B$ blocks, each block may choose the number of layers from $\mathbf{L} = \{1, 2,3,4,5\}$, a filter length from $\mathbf{F}=\{10,20,40,80,160\}$, and a bit-width 
from $\mathbf{W}=\{4,8,16,32\}$. 
The selected filter length applies to the first layer. In the remaining layers, the filter lengths are reduced by half. 
The search space then includes
%
${(\mathbf{|L|} \times \mathbf{|W|} \times \mathbf{|F|})^B}$
different student settings, which amounts to a very large space.

A student setting $\mathbf{x}_i$ in the search space is a sequence of $B$ entries, corresponding to $B$ blocks. For the $j$-th entry, $1\leq j\leq B$, tuple $(L_j, F_j, W_j)$ indicates that there are $L_j$ layers, that the filter length in the first layer is $F_j$, and that the bit-width is $W_j$, 
as shown in Equation~\ref{eq:setting}. 
%
For example, with $B=3$ blocks, the student setting $\mathbf{x}_0$ has three entries, where the first entry $(3, 20, 8)$ indicates that there are $3$ layers in the first block, that the filter length of the first layer is $20$, meaning that the filter lengths of the following layers are 10 and 5, and that the bit-width of filters is 8 bits. 
\begin{gather}
    \mathbf{x}_i = 
    \begin{pmatrix} 
	(L_1, F_1, W_1)\\
	(L_2, F_2, W_2)\\
	\dots \\
	(L_B, F_B, W_B)\\
	\end{pmatrix},
	\ \mathbf{x}_0 = 
	\begin{pmatrix} 
	(3, 20, 8) \\
	(4, 40, 4) \\
	(2, 10, 16) \\
	\end{pmatrix} 
	\label{eq:setting}
\end{gather}

\noindent
\textbf{Applicability to Other Neural Classifiers: } 
Although we use \texttt{InceptionTime} as the base model, the proposed method is not limited to \texttt{InceptionTime}. With minor adjustments, the search space can be adapted to accommodate 
a variety of neural classifiers. 
More specifically, the choices of the numbers of blocks and layers and the bit-widths apply universally across different neural classifiers. The filter lengths are specific to convolutional classifiers such as \texttt{InceptionTime}. When using other types of classifiers, other components need to be considered. 
%
%
For fully-connected classifiers~\cite{WangYO17}, the number of neurons per layer needs to be included into the search space; 
for recurrent neural network based classifiers~\cite{KarimMD19}, the sizes of their recurrent weight matrices need to be included into the search space; 
%
for Transformer\Hyphdash based classifiers~\cite{ZhangG0L20}, the sizes of the projection matrices need to be included. 
%
Thus, the search space
can be adapted easily to different classifiers. In this paper, we consider a search space based on \texttt{InceptionTime} because it outperforms other classifiers at time series classification~\cite{RuizFLMB21}. 

\subsubsection{Pareto Frontier}

Given a student setting, we can obtain its accuracy using the proposed \texttt{AED}, but this is costly. 
%
Next, it is also possible to compute the size based on the setting using the information on the bits per layer, the number of filters, and filter lengths (see Section~\ref{sec:quantized}), which is very efficient. 
%
Thus, for each setting, we can define a tuple $s$ that includes its associated accuracy and its model size---see Equation~\ref{eq:tuple}.
%
\begin{gather}
    s_i = (\mathbf{x}_i, {\mathit accuracy_i}, {\mathit size_i})
    \label{eq:tuple}
\end{gather}
We use $\mathbf{S}$ to denote the set of all student settings. Given two settings $s_1,s_2 \in \mathbf{S}$, $s_2$ dominates $s_1$, denoted as $s_2 \succ s_1$, if $s_2$ is more accurate and not larger than $s_1$ or if $s_2$  is smaller than $s_1$ and is not worse than $s_1$ in terms of accuracy. 

The Pareto frontier $P(\mathbf{S})$ is a subset of $\mathbf{S}$, where for every tuple $s \in P(\mathbf{S})$ no other tuple $s^\prime$ exists that dominates $s$, as shown in Equation~\ref{eq:pareto_frontier}.
\vspace*{-0.3em}
\begin{align}
    P(\mathbf{S}) = \{s \in \mathbf{S} \, | \, \nexists  \{s^\prime \in \mathbf{S} \mid s^\prime \succ s\wedge s^\prime \ne s\}\}
    \label{eq:pareto_frontier}
\end{align}

The frontier facilitates the identification of the setting with the highest accuracy under specific model size constraints. Thus, the aim is to identify the Pareto frontier. 

\subsubsection{Encoded Multi-objective Bayesian Optimization}
\label{sssec:EnMOBO}

To get the exact Pareto frontier, it is necessary to perform domination comparisons on the ${(\mathbf{|L|} \times \mathbf{|W|} \times \mathbf{|F|})^B}$ settings in the search space, e.g., using skyline querying algorithms~\cite{BorzsonyiKS01}. 
%
%
However, since 
evaluating the accuracy for a setting using \texttt{AED} is costly, it is infeasible to evaluate the accuracy of all ${(\mathbf{|L|} \times \mathbf{|W|} \times \mathbf{|F|})^B}$ settings using \texttt{AED}.  
Instead, we 
evaluate the accuracy of a small subset of $Q$ ($Q\ll{(\mathbf{|L|} \times \mathbf{|W|} \times \mathbf{|F|})^B}$) settings, to obtain an approximate frontier. 
%

\begin{figure}[ht!]
    \centering
    \includegraphics[width=0.93\linewidth]{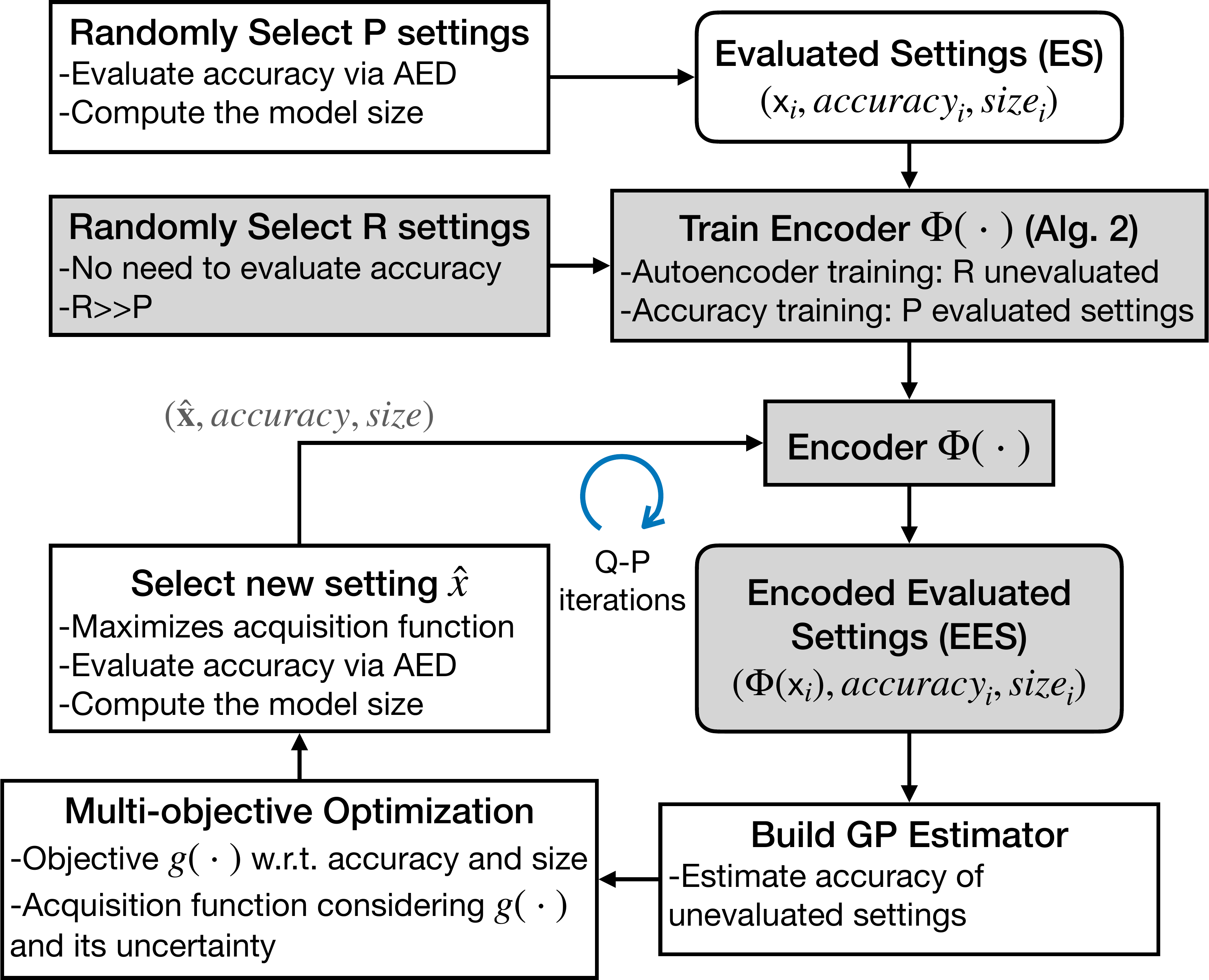}
    \caption{Encoded Multi-Objective Bayesian Optimization.}
    \label{fig:Bayesian}
\end{figure}

A naive way to select the $Q$ settings is to apply random sampling. 
Although being simple, random sampling may not be effective as it takes into account neither accuracy nor model size. 
We propose instead a novel encoded multi-objective Bayesian optimization method to focus on exploring the most promising settings in the search space, by considering both accuracy and model size---see Figure~\ref{fig:Bayesian}.

To initialize the optimization process, we first randomly select $P$ settings, $P<Q$.
We evaluate these using \texttt{AED} to obtain their accuracies and also compute their sizes. 
%
%
They constitute the set of evaluated settings ($ES$).

In classic multi-objective Bayesian optimization (MOBO), as shown in the white boxes in Figure~\ref{fig:Bayesian}, the evaluated settings $ES$ are used to build a Gaussian Process (GP) that estimates the accuracies of the unevaluated settings.
We use a GP because it is able to estimate not only the accuracy but also a
probability distribution of the estimated accuracy, which is helpful for guiding the subsequent exploration. 
%
An objective function is built considering both the estimated accuracies and model sizes, and 
then an acquisition function helps identify the most promising setting $\hat{\mathbf{x}}$ based on the objective function. We evaluate $\hat{\mathbf{x}}$'s accuracy using \texttt{AED}, 
compute its size, and feed it back to the 
GP estimator to get it updated. Then, a new iteration starts until we have evaluated $Q$ settings. 

\vspace{0.2cm}
\noindent
\textbf{GP estimator: }
A GP is a probabilistic regression model, that learns a mapping $f(\cdot)$ from a multidimensional point, e.g., a student setting $\mathbf{x}_i$, to a real value, e.g., the accuracy $\mu_i$ of the setting, along its probability distribution, so 
$f(\mathbf{x}_i) \sim \mathcal{N}(\mu_i,\,\sigma_i^{2})$.
Given $P$ evaluated settings $\mathbf{x}_{1:P}$, 
the joint distribution is: 
\begin{equation} \label{eq:joint_dist}
\mathbf{f}(\mathbf{x}_{1:P})\sim \mathcal{N}(\texttt{AED}(\mathbf{x}_{1:P}),\mathbf{K}(\mathbf{x}_{1:P},\mathbf{x}_{1:P})),
\end{equation}
where \texttt{AED}($\mathbf{x}_{1:P}$) is the evaluated accuracies for the $P$ settings, and $\mathbf{K}$ is the covariance matrix with kernel function $\kappa(\mathbf{x}_i,\mathbf{x}_j)$.  
A generally used kernel $\kappa$ is the 
squared exponential covariance function: $\kappa(\mathbf{x}_{i},\mathbf{x}_{j})=\theta_{f} \mathbf{exp}\left (  -\left | \frac{(\mathbf{x}_{i}-\mathbf{x}_{j})^{2}}{2\Theta ^{2}} \right |\right )$ with variance $\theta_{f}$, scaled by the noise level $\Theta$ of the observations.

Taking the GP as the prior~\cite{rasmussen-et-al:gp}, the posterior predictive distribution for an unevaluated setting $\mathbf{x}^*$ is defined as:
\begin{equation} \label{eq:gp}
f(\mathbf{x}^*)\sim \boldsymbol{N}({\mu}(\mathbf{x}^*),\sigma^2(\mathbf{x}^*)),
\end{equation}
where $\mu(\mathbf{x}^*)=\kappa(\mathbf{x}^*,\mathbf{x}_{1:P})\mathbf{K}(\mathbf{x}_{1:P},\mathbf{x}_{1:P})^{-1}\mathbf{f}(\mathbf{x}_{1:P})$, and $\sigma^2(\mathbf{x}^*)= \kappa(\mathbf{x}^*,\mathbf{x}^*)-\kappa(\mathbf{x}^*,\mathbf{x}_{1:P})$ $\mathbf {K}(\mathbf{x}_{1:P},\mathbf{x}_{1:P})^{-1}\kappa (\mathbf{x}^*,\mathbf{x}_{1:P}) ^{T}$.

As the kernel function $\kappa(\mathbf{x}_{i},\mathbf{x}_{j})$ suggests, the distance between two different settings affects the GP estimator significantly: if the distance between two settings is small, the are expected to have similar accuracy. Therefore, a meaningful distance metric between settings is important. 
We proceed to illustrate why the Euclidean distance on the space of original settings fails to be meaningful and then elaborate a novel encoding scheme such that the Euclidean distance on the encoded space offers meaningful distances. 

\noindent
\textit{Complexity: }
The process shown in Figure~\ref{fig:Bayesian} requires running \texttt{AED} with teacher removal $Q$ times, yielding $\mathcal{O}(Q\times N\times E \times BP_w)$. 
In addition, the GP estimator is run $Q-P$ times. 
The complexity of a GP estimator is $\mathcal{O}(n^3)$ due to the matrix inverse operation, where $n$ is the size of the covariance matrix $\mathbf{K}$. The size of $\mathbf{K}$ keeps increasing as the steps continues and reaches $Q$ in the last step. 
Thus, the GP part is $\mathcal{O}(Q^4)$. In total, we get $\mathcal{O}(Q\times N\times E \times BP_w+Q^4)$.

\vspace{0.2cm}
\noindent
\textbf{Problems of the Original Space: }
Consider three settings $\mathbf{x}_1$, $\mathbf{x}_2$, and $\mathbf{x}_3$, all with three blocks, as shown in Equation~\ref{eq:simplified_settings}. We also show their accuracies and the Euclidean distances $\overline{\mathbf{x}_1 \mathbf{x}_2}$ and $\overline{\mathbf{x}_1 \mathbf{x}_3}$ in the original space. 
\begin{gather}
	\begin{matrix}
	\mathbf{x}_1 = 
	\begin{pmatrix} 
	(4,40,8) \\
	(4,40,8) \\
	(4,40,8) \\
	\end{pmatrix} ,\\
    \hspace{-8ex} \text{Accuracy:} \quad 0.37 \\
	\end{matrix} \,
	\begin{matrix}
	\mathbf{x}_2 = 
	\begin{pmatrix} 
	(1,40,8) \\
	(1,40,8) \\
	(1,40,8) \\
	\end{pmatrix} ,\\
    \qquad 0.24
	\end{matrix} \,
	\begin{matrix}
	\mathbf{x}_3 = 
	\begin{pmatrix} 
	(4,40,16) \\
	(4,40,16) \\
	(4,40,16) \\
	\end{pmatrix} \\
    \qquad 0.38
	\end{matrix}
	\label{eq:simplified_settings} \\
	\text{Distances} \quad
	\begin{matrix*}[l]
	\text{Original} & \overline{\mathbf{x}_1 \mathbf{x}_2}=5.19 & \overline{\mathbf{x}_1 \mathbf{x}_3}=13.85 \\
	\text{Encoded} & \overline{\mathbf{x}_1 \mathbf{x}_2}=3.33 & \overline{\mathbf{x}_1 \mathbf{x}_3}=1.70\\
	\end{matrix*}
	\label{eq:distances}
	\nonumber
\end{gather}
Settings $\mathbf{x}_1$ and $\mathbf{x}_2$ have the same bit-width, but they have different numbers of layers, $4$ vs. $1$, yielding quite different neural structures and dissimilar accuracies. In contrast, $\mathbf{x}_1$ and $\mathbf{x}_3$ differ on the bit-width, $8$ vs. $16$, but share the same number of layers and have similar accuracy. Based on the above, $\overline{\mathbf{x}_1 \mathbf{x}_3}$ should be smaller than $\overline{\mathbf{x}_1 \mathbf{x}_2}$, as $\mathbf{x}_1$ and $\mathbf{x}_3$ have more similar accuracies.  

However, in the original space, $\overline{\mathbf{x}_1 \mathbf{x}_3}$ is much larger than $\overline{\mathbf{x}_1 \mathbf{x}_2}$, as the bit-width difference $(16-8)^2$ is larger than the layer difference $(4-1)^2$. Thus, the distance in the original space is not aligned with the accuracy similarity. 
This happens because the values are discrete and the different semantics of the different values are also different dimensions, so
adding up the distances directly is semantically incorrect.

\noindent
\textbf{A Novel Encoding Scheme: }
To address the problem of inconsistent distances, we propose to utilize a continuous latent space instead of the original discrete values to capture semantic similarities. In addition, a continuous space is known to fit GPs better~\cite{DeshwalD21,TrippDH20}. Thus, we propose an encoder $\Phi: \mathbf{x} \rightarrow \mathbf{z}$ that maps the setting $\mathbf{x_i}$ to a continuous feature vector $\mathbf{z_i}$. Then, we use the encoded $\mathbf{z_i}$ as the input to the GP. 

\begin{figure}[ht!]
    \centering
    \includegraphics[width=0.8\linewidth]{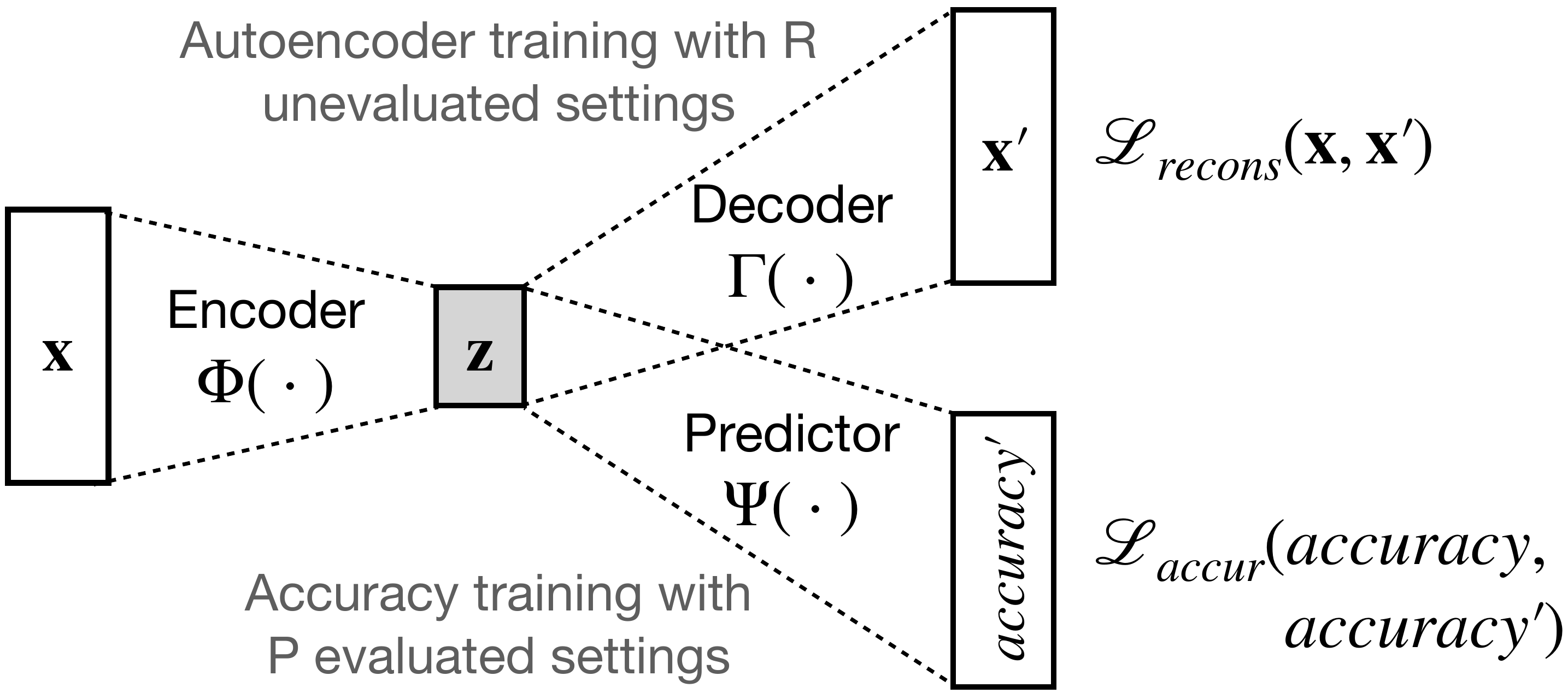}
    \caption{Two-phase Training for Encoder $\Phi(\cdot)$.}
    \label{fig:autoencoder}
\end{figure}

We proceed to cover the training of the encoder $\Phi$. The training involves two phases, as illustrated in Figure~\ref{fig:autoencoder} and detailed in Algorithm~\ref{alg:autoencoder}. 
In the first phase, we connect the encoder with a decoder $\Gamma(\cdot)$. To train the encoder and decoder jointly, we random sample $R$ unevaluated settings, without knowing the accuracies of the $R$ settings, where $R>>P$. We use the decoder to reconstruct the $R$ settings
$\{\mathbf{x_r}\}_{r=1}^R$. 
It computes $\mathbf{x}^{\prime}_r = \Gamma(\Phi(\mathbf{x_r}))$ and then minimizes the loss $\mathcal{L}_{{\mathit recons}} (\mathbf{x_r}, \mathbf{x}_r^{\prime}) = \frac{1}{R} \sum_{r=1}^R (\mathbf{x_r} - \mathbf{x}_r^{\prime})^2$. 
%
%
In other words, the encoder and decoder work together as an autoencoder. This enables an encoder that is able to transform a discrete setting $\mathbf{x_r}$ into a continuous vector $\mathbf{z_r}$. However, since the training data remains ``accuracy blind,'' the encoded continuous space fails to capture the semantics related to accuracy. 
In the second phase, 
we therefore connect the encoder with an accuracy predictor $\Psi(\cdot)$ and train the encoder and predictor jointly using the $P$ evaluated settings $\{\mathbf{x_p}\}_{p=1}^P$, so the predictor estimates ${\mathit accuracy_p}^{\prime} = \Psi(\Phi(\mathbf{x_p}))$ by minimizing the loss $\mathcal{L}_{{\mathit accur}} ({\mathit accur_p, accur_p}^{\prime}) = \frac{1}{P} \sum_{p=1}^P ({\mathit accur_p - accur_p}^{\prime})^2$. Since the encoder is already trained with the decoder using $R$ unevaluated settings, we use an only small amount of $P$, where $P<<R$, evaluated settings to fine tune it, such that the encoded space is aligned better with accuracy. 
As Algorithm~\ref{alg:autoencoder} shows, we use the two training phases interchangeably. 

\algblock{Input}{EndInput}
\algnotext{EndInput}
\algblock{Output}{EndOutput}
\algnotext{EndOutput}
\begin{algorithm}
\caption{Two-phase Encoder $\Phi(\cdot)$ Training.}
\label{alg:autoencoder}
\begin{algorithmic}[1]

\Input : R unevaluated settings: $\{\mathbf{x_r}\}$, P evaluated settings: $(\mathbf{x_i}, {\mathit accuracy_i})$
\EndInput
\Output : Encoder $\Phi$
\EndOutput
\State $ps \gets \text{Predictor steps}$
\State Encoder $\Phi$, Decoder $\Gamma$, Predictor $\Psi$
\For {$e \gets 1,2,\ldots, Epochs$}
    \State $\mathbf{x}_r^{\prime} \gets \Gamma(\Phi(\mathbf{x_r}))$ \Comment{Autoencoder training, using $R$}
    \State $\arg \min_{\Phi, \Gamma} \, \mathcal{L}_{{\mathit recons}} (\mathbf{x_r}, \mathbf{x}_r^{\prime})$
    \If {$e \mod ps = 0$} \Comment{Predictor training, using $P$}
        \State ${\mathit accuracy_p^{\prime}} \gets \Psi(\Phi(\mathbf{x_p}))$
        \State $\arg \min_{\Phi, \Psi} \, \mathcal{L}_{{\mathit accur}} {\mathit (accuracy_p, accuracy_p^{\prime})}$
    \EndIf
\EndFor
\end{algorithmic}
\end{algorithm}

%
The integration of the encoding scheme into the overall Bayesian Optimization process is shown in Figure~\ref{fig:Bayesian} using grey boxes. After training the Encoder $\Phi(\cdot)$ using $R$ unevaluated settings and $P$ evaluated settings, it is used to map all the evaluated settings ($ES$) to the corresponding encoded space, getting the Encoded Evaluated Settings ($EES$). Then, based on the $EES$,
we build a GP to estimate the accuracy of the unevaluated settings. 
%
To facilitate the identification of settings on the Pareto frontier, given an unevaluated setting $\mathbf{x}^*$, we define a joint objective function $g(\mathbf{x}^*) = \beta \times f(\mathbf{x}^*) - (1-\beta) \times {\mathit Size}(\mathbf{x}^*)$ that considers both the estimated accuracy $f(\mathbf{x}^*)$ and the computed model size ${\mathit Size}(\mathbf{x}^*)$, preventing divergent optimizations if they are considered separately. We use different $\beta$ in different optimization iterations, facilitating searches with different trade-offs between accuracy and model size. More specifically, $\beta$ is randomly sampled following the PACE model strategy~\cite{JonesSW98}. Since the estimated accuracy $f(\mathbf{x}^*)$ is uncertain, the joint objective $g(\mathbf{x}^*)$ is also uncertain. 
Next, we use the Expected Improvement ($EI$) acquisition function~\cite{DaultonBB20,DaultonBB21} based on the joint objective $g(\cdot)$, to determine the most promising setting $\hat{\mathbf{x}}$. 
We evaluate $\hat{\mathbf{x}}$'s accuracy using \texttt{AED} and compute its size, and we use the trained encoder $\Phi(\cdot)$ to encode $\hat{\mathbf{x}}$ and then add it to $EES$. Based on the updated $EES$, the GP estimator is also updated. Then, a new iteration starts.
\section{Experiments} \label{sec:experiments}

\subsection{Experimental Setup}

\subsubsection{Data Sets}
We use the UCR time series archive~\cite{DauBKYZGRK19} that comprises an extensive catalogue of time series from different domains. 
Experiments are conducted on the 128 available data sets, and we focus on data sets with many classes since they are the most challenging and have important real-world uses. 
%
Details of the selected subset of the data sets are shown in Table~\ref{table:data_sets}, including the number of classes, the sizes of the training, validation, and testing sets, the domains, and the average time series lengths.

\begin{table}[ht!]
    \small
    \centering
    \caption{Data Sets.}
    \label{table:data_sets}
    \begin{tabular}{ |l|c|c|c|c| } 
    \hline
    \textbf{Data set} &  \textit{Classes} &  \textit{Train/Val/Test} &  \textit{Domain} &  \textit{Avg. Len.}\\ 
    \hhline{|-|-|-|-|-|}
    
    \textit{Adiac} & 37 & 312/78/391 & Images & 176  \\
    \textit{Crop} & 27 & 5720/1440/16800 & Images & 46 \\
    \textit{FaceAll} & 14 & 448/112/1690 & Images & 131  \\
    \textit{NonInvECG1} & 42 & 1440/360/1965 & ECG & 750  \\
    \textit{NonInvECG2} & 42 & 1440/360/1965 & ECG & 750 \\
    \textit{Phoneme} & 39 & 171/43/1896 & Sound & 1024 \\
    \textit{PigAirway} & 52 & 83/19/208 & Blood flow & 2000 \\
    \textit{PigArt} & 52 & 83/19/208 & Blood flow & 2000 \\
    \textit{UWave} & 8 & 1680/560/2241 & Motion & 315 \\
    \hline
    \end{tabular}
\end{table}

\subsubsection{Evaluation Metrics}
To assess performance, we consider several metrics. First, \textit{Accuracy} is the percentage of cases where the  
class with the highest output probability has 
the correct class label. 
It 
is also used in conjunction with the null-hypothesis Friedman test~\cite{Friedman1940ACO} and the Wilcoxon-Holm post-hoc method~\cite{Wilcoxon,Holm} to evaluate  all data sets. Next, \textit{Top-5 Accuracy} evaluates the classification results based on the top-5 output probabilities. If the probability associated with the correct label is within the top-5, the result is considered as correct. It is used in the experiments involving data sets with many labels. Finally, \emph{Model size} indicates how much memory a model requires for a specific configuration of layers and bit-width. 

\subsubsection{Baselines}
\texttt{LightTS} is compared to four knowledge distillation methods. Although these methods use different strategies to combine the results from base models into a single teacher model, they share the same distillation scheme---the distillation is conducted between a single teacher and a student. (1) In Classic Knowledge Distillation (\texttt{Classic KD})~\cite{HintonVD15,YangSGLJ20}, the teacher is the average of the class probabilities across all base models, as shown in Equation~\ref{eq:adjusted_loss} in Section~\ref{sec:knowledge_other}. The next three baselines aims at improving \texttt{Classic KD} by proposing different techniques to weigh the base models differently, i.e., finding values other than the $1/N$ in Figure~\ref{subfig:knowledge}. (2) Adaptive Ensemble (\texttt{AE-KD})~\cite{DuYLW00Z20} 
optimizes the weights using Support Vector Machines according to the teachers diversity derived from gradient changes. (3) Reinforced Multi-Teacher (\texttt{Reinforced})~\cite{YuanSPLGFJ21} computes the weights following a reinforced learning process using the training losses as the reward metric. (4) Cross-validation Accuracy Weighted Probabilistic Ensemble (\texttt{CAWPE})~\cite{LargeLB19} uses the cross-validation accuracy results to assign the weights. Thus, if a model has a high accuracy during validation, it has a high weight in the final ensemble. In addition, we include two variations of \texttt{LightTS}. (5) \texttt{AED-LOO}: we use \textit{AED} with leave-one-out teacher removal. (6) \texttt{AED-One}: we use \textit{AED} without teacher removal, i.e., 
Algorithm~\ref{alg:ensemble_selection} is run only once. 
Finally, we also include the full-precision ensemble \texttt{FP-Ensem} for reference. 
As knowledge distillation can boost student performance~\cite{YimJBK17,TianKI20}, a quantized student can possibly outperform \texttt{FP-Ensem}. 

\subsubsection{Ensemble Teachers}
To evaluate the generality of \texttt{LightTS} on different types of ensembles, we consider ensembles with a wide variety of base models.
The ensembles for all cases have $N=10$ base models, and the base models are initialized with different random states to ensure diversity.
The base models used are detailed as follows. (1) InceptionTime~\cite{FawazLFPSWWIMP20} is the state-of-the-art deep learning method~\cite{RuizFLMB21}. 
It is the default base model in the experiments. We consider three types of non-deep learning based methods that are components for state-of-the-art meta-ensemble classifiers~\cite{MiddlehurstLFLB21}. (2) Temporal Dictionary Ensemble (\textit{TDE})~\cite{MiddlehurstVB19} transforms a time series into a bag of segments of a given size and discretizes them as words. Then, it draws a histogram for the word counting and applying a nearest neighbor algorithm to classify the transformed series. (3) Canonical Interval Forest Classifier (\textit{CIF})~\cite{MiddlehurstLB20} builds a Time Series Forest using a set of 22 particular features~\cite{LubbaSKSFJ19} to summarize them in intervals that are used to classify the time series. (4) Time Series Forest Classifier (\textit{Forest})~\cite{DengRTM13} builds several trees for representing a time series, summarizing them in intervals. Then, a forest is built over the trees to identify singular features in the intervals to perform the classification of the series. 

\subsubsection{Implementation Details} \label{sss:implementation}
The \texttt{LightTS} framework is implemented using Python 3.9.7, the machine learning architecture PyTorch 1.9.1, and the Bayesian Optimization infrastructure of BoTorch 0.5.1. 
%
The source code is publicly available at 
{\url{https://github.com/d-gcc/Distiller}}.
All the models are tested using Titan RTX GPUs with 24GB of VRAM under Ubuntu 20.04.3 on an Intel Xeon W-2155 with 128GB of RAM.

To ensure fair comparisons, we follow common machine learning practices and use the validation set to adjust the hyper-parameters for all methods. 
The ensemble of ten base teacher models is trained in full\Hyphdash precision using different random seeds to ensure diversity. 
They are trained for 1,500 epochs, with a learning rate of 0.01, Adam optimized, and using a batch size of 64. 
Then, following the same configuration, the distilled student is quantized according to the testing configuration, uses a validation set of 20\% with Stochastic Gradient Descent as the optimizer, and the weights $\boldsymbol\lambda$ are adjusted every $v=50$ epochs. 
The predictor for the encoded MOBO is also adjusted every $50$ epochs. 

\subsection{Experimental Results} \label{subsec:exp_results}
%

\subsubsection{Problem Scenario 1} \label{subsub:problem1}
We consider lightweight students with 3 blocks and 3 layers per block. All layers use the same bit-width, chosen among 4-bit, 8-bit, and 16-bit. This means that the lightweight students obtained by different methods have the same size. Thus, we only evaluate their accuracy.

\begin{figure}[ht]
    \vspace*{-1em}
    \centering
    \includegraphics[width=\linewidth]{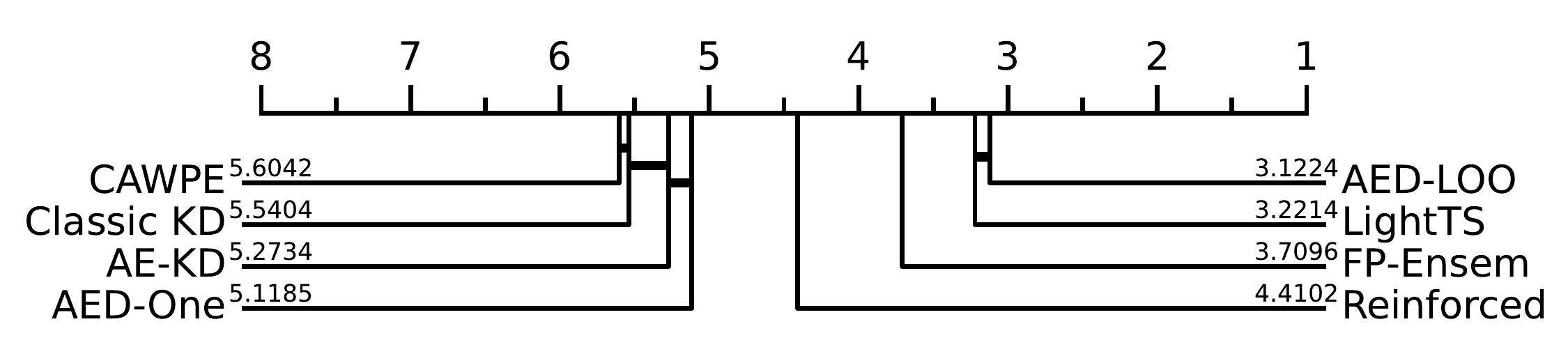}
    \vspace*{-2.5em}
    \caption{\textit{Accuracy} Ranking, Full UCR Archive.}
    \label{fig:all-bit}
\end{figure}

\noindent
\textbf{\texttt{InceptionTime} as Base Models:}
We report the results for the complete UCR data sets using \texttt{InceptionTime} as the base models. 
Figure~\ref{fig:all-bit} shows the critical difference diagram on {\textit{Accuracy}} after applying the null-hypothesis Friedman test and the Wilcoxon-Holm post-hoc test to rank the evaluated methods. The evaluated methods are ranked according to the pairwise comparison of accuracy for every set and bit-width. Then, the average rank across all the data sets and all bit-width settings is computed, as the diagram shows. 

\begin{figure}[ht]
    \vspace*{-1em}
    \centering
    \includegraphics[width=\linewidth]{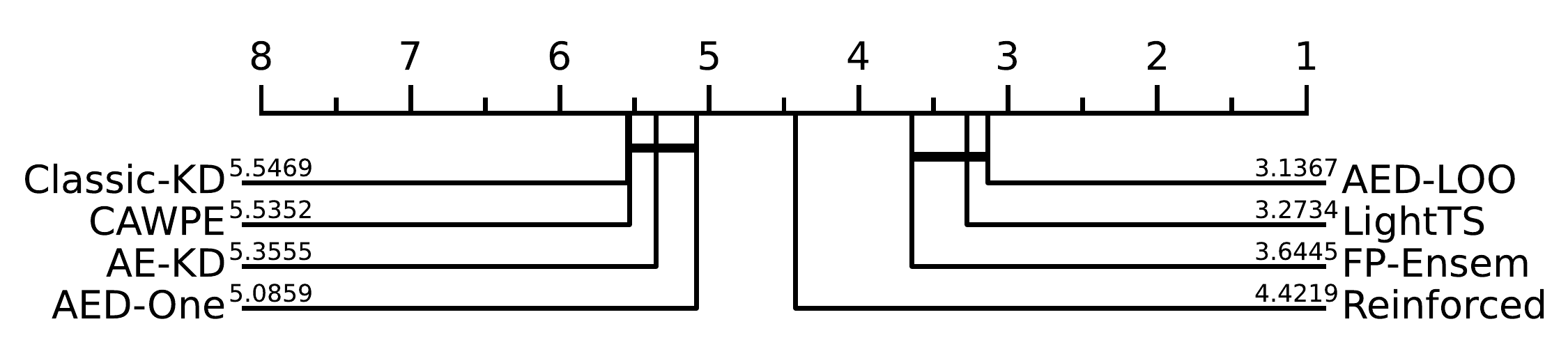}
    \vspace*{-2em}
    \caption{4-bit \textit{Accuracy} Ranking.}
    \label{fig:rank4-bit}
    \vspace*{-1em}
\end{figure}

\begin{figure}[ht]
    \vspace*{-0.5em}
    \centering
    \includegraphics[width=\linewidth]{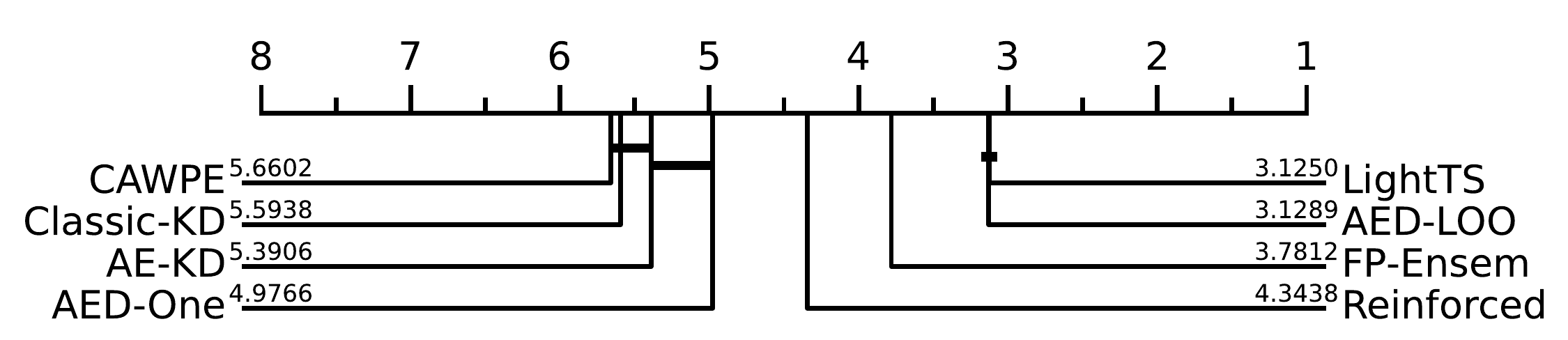}
    \vspace*{-2em}
    \caption{8-bit \textit{Accuracy} Ranking.}
    \label{fig:rank8-bit}
    \vspace*{-1em}
\end{figure}

\begin{figure}[ht]
    \vspace*{-0.5em}
    \centering
    \includegraphics[width=\linewidth]{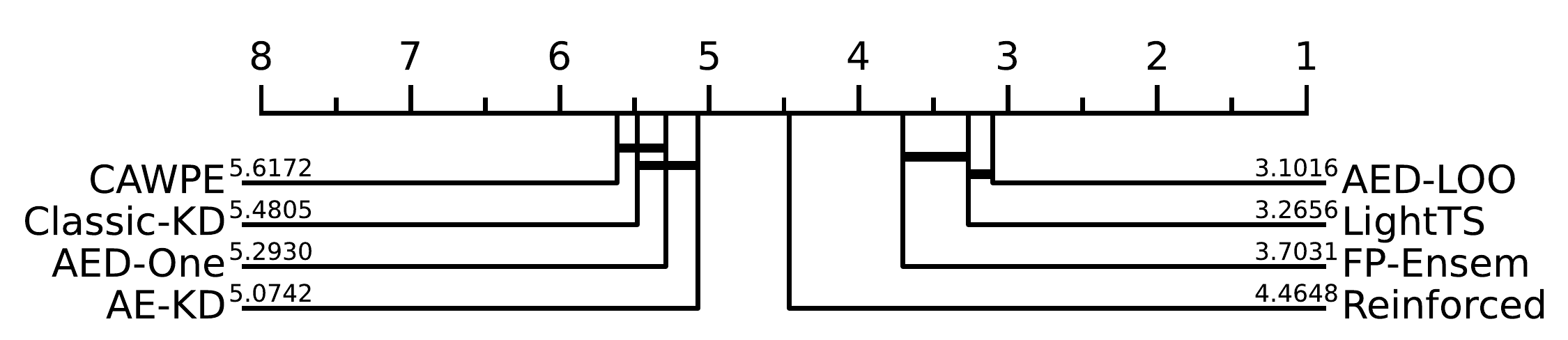}
    \vspace*{-2em}
    \caption{16-bit \textit{Accuracy} Ranking.}
    \label{fig:rank16-bit}
\end{figure}

Figure~\ref{fig:all-bit} shows that our proposals \texttt{LightTS} and \texttt{AED-LOO} achieve the best results. They are clustered in the first rank, and they are statistically more accurate than the other methods. 
In particular, \texttt{LightTS} and \texttt{AED-LOO} are statistically more accurate than the full-precision ensemble \texttt{FP-Ensem}, which is ranked in second. 
This may sound counter-intuitive, but this is possible because knowledge distillation can boost student performance, enabling a well-distilled quantized student to outperform \texttt{FP-Ensem}.  
Then comes \texttt{Reinforced}, which is followed by a statistically similar cluster of the remaining baselines led by \texttt{AED-One}. 
Disaggregated results for 4-bit, 8-bit, and 16-bit are shown in Figures~\ref{fig:rank4-bit}, \ref{fig:rank8-bit}, and \ref{fig:rank16-bit}, respectively.

\begin{figure}[ht]
    \vspace*{-1em}
    \centering
    \includegraphics[width=\linewidth]{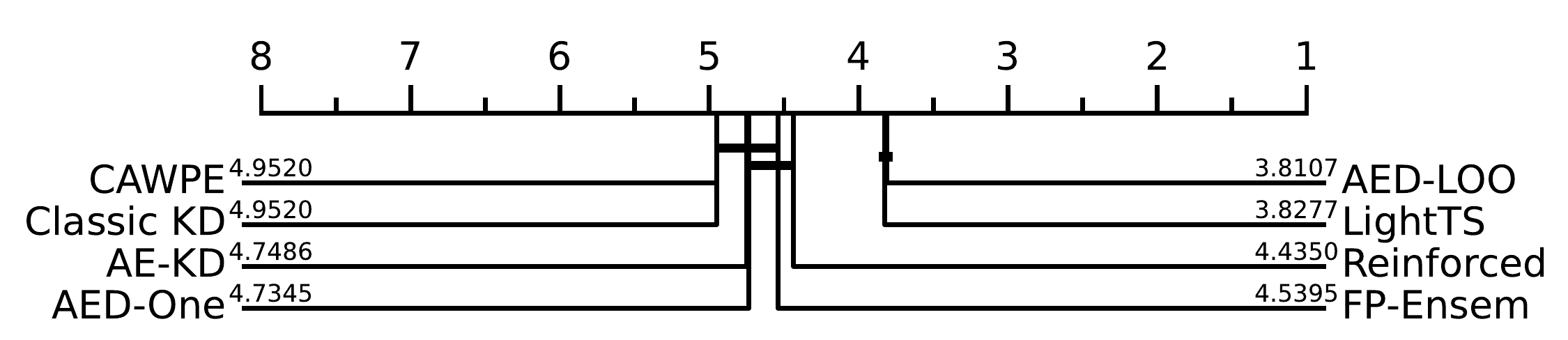}
    \vspace*{-2.5em}
    \caption{\textit{Accuracy} Ranking, 59 Data Sets with 2 or 3 Classes.}
    \label{fig:rank-all-fewclasses}
\end{figure}

For data sets with 2 or 3 classes, representing 46\% of the UCR data sets, we explore the results in Figure~\ref{fig:rank-all-fewclasses}. 
It shows a similar ranking as in the overall results, with all the methods having closer results, which is expected as several pairwise comparisons are tied giving the reduced number of classes. This suggests that \texttt{LightTS} works equally well on data sets with few classes. 

Detailed results for challenging data sets with many labels are shown in Table~\ref{table:Main}. For those cases,
in addition to the baselines, 
we consider a full-precision student, denoted as \texttt{FP-Stud}. This is a student model with 32-bit model parameters, which based on knowledge distilled from the full precision ensemble using \texttt{AED}. \texttt{FP-Stud} is supposed to offer an accuracy upper bound to the quantized student with 4-bit, 8-bit, and 16-bit model parameters, denoted by a rectangle, e.g., the fourth row of Table~\ref{table:Main}.
%

\addtolength{\tabcolsep}{-1pt}
\begin{table*}[ht!]
\setlength\fboxsep{1pt}
    \small
    \centering
    \caption{Accuracy of Lightweight Student Models, using an Ensemble of \texttt{InceptionTime} Base Models.}
    \label{table:Main}
    \begin{tabular}{ |l|c|c|c|c|c|c|c|c|c|c|c|c|c|c|c|c|c|c| } 
    \hline

    \textbf{} & \multicolumn{3}{c|}{\textit{Accuracy}} & \multicolumn{3}{c|}{\textit{Top-5 Accuracy}} & \multicolumn{3}{c|}{\textit{Accuracy}} & \multicolumn{3}{c|}{\textit{Top-5 Accuracy}} & \multicolumn{3}{c|}{\textit{Accuracy}} & \multicolumn{3}{c|}{\textit{Top-5 Accuracy}} \\ 
    \hline
    \textbf{Bit-width} & \textit{4} & \textit{8} & \textit{16} & \textit{4} & \textit{8} & \textit{16}
    & \textit{4} & \textit{8} & \textit{16} & \textit{4} & \textit{8} & \textit{16}
     & \textit{4} & \textit{8} & \textit{16} & \textit{4} & \textit{8} & \textit{16}
    \\ 
    \hline
\textbf{} & \multicolumn{6}{l|}{\textit{Adiac}} & \multicolumn{6}{l|}{\textit{NonInvECG1}} & \multicolumn{6}{l|}{\textit{PigAirway}}\\
\hline
\texttt{FP-Ensem/FP-Stud} & \multicolumn{3}{c|}{0.79 / \boxed{0.83}} & \multicolumn{3}{c|}{0.95 / \boxed{0.97}} & \multicolumn{3}{c|}{0.96 / \boxed{0.96}} & \multicolumn{3}{c|}{1.00 / \boxed{1.00}} & \multicolumn{3}{c|}{0.56 / \boxed{0.59}} & \multicolumn{3}{c|}{0.91 / \boxed{0.93}}\\
\hhline{|~|-|-|-|-|-|-|-|-|-|-|-|-|-|-|-|-|-|-|}
\texttt{Classic KD} & 0.29 & 0.41 & 0.48 & 0.65 & 0.80 & 0.83 & 0.55 & 0.68 & 0.78 & 0.91 & 0.94 & 0.99 & 0.18 & 0.24 & 0.27 & 0.32 & 0.50 & 0.63 \\
\texttt{AE-KD} & 0.29 & 0.36 & 0.38 & 0.51 & 0.56 & 0.74 & 0.57 & 0.63 & 0.78 & 0.98 & 0.98 & 0.99 & 0.19 & 0.22 & 0.24 & 0.42 & 0.55 & 0.55 \\
\texttt{Reinforced} & 0.20 & 0.23 & 0.40 & 0.41 & 0.51 & 0.87 & 0.56 & 0.62 & 0.78 & 0.92 & 0.95 & 0.99 & 0.14 & 0.21 & 0.26 & 0.54 & 0.57 & 0.59 \\
\texttt{CAWPE} & 0.23 & 0.27 & 0.38 & 0.48 & 0.56 & 0.82 & 0.72 & 0.77 & 0.77 & 0.97 & 0.99 & 0.99 & 0.17 & 0.21 & 0.26 & 0.44 & 0.48 & 0.68 \\
\hline
\texttt{AED-LOO} & 0.76 & \textbf{0.78} & 0.78 & 0.96 & \textbf{0.97} & \textbf{0.97} & \textbf{0.94} & \textbf{0.95} & 0.94 & \textbf{1.00} & \textbf{1.00} & \textbf{1.00} & \textbf{0.56} & \textbf{0.57} & \textbf{0.59} & \textbf{0.91} & \textbf{0.91} & \textbf{0.92} \\
\texttt{LightTS} & \textbf{0.77} & 0.77 & \textbf{0.79} & \textbf{0.97} & \textbf{0.97} & \textbf{0.97} & 0.92 & 0.94 & \textbf{0.95} & \textbf{1.00} & \textbf{1.00} & \textbf{1.00} & 0.53 & 0.54 & 0.55 & 0.90 & 0.90 & 0.90 \\

\hline
\textbf{} & \multicolumn{6}{l|}{\textit{Crop}} & \multicolumn{6}{l|}{\textit{NonInvECG2}} & \multicolumn{6}{l|}{\textit{PigArt}}\\
\hline
\texttt{FP-Ensem/FP-Stud} & \multicolumn{3}{c|}{0.76 / \boxed{0.76}} & \multicolumn{3}{c|}{0.95 / \boxed{0.97}} & \multicolumn{3}{c|}{0.96 / \boxed{0.96}} & \multicolumn{3}{c|}{1.00 / \boxed{1.00}} & \multicolumn{3}{c|}{0.99 / \boxed{1.00}} & \multicolumn{3}{c|}{0.99 / \boxed{1.00}} \\
\hhline{|~|-|-|-|-|-|-|-|-|-|-|-|-|-|-|-|-|-|-|}
\texttt{Classic KD} & 0.67 & 0.68 & 0.69 & 0.96 & 0.96 & 0.96 & 0.80 & 0.81 & 0.83 & 0.99 & 0.99 & 0.99 & 0.53 & 0.63 & 0.67 & 0.64 & 0.89 & 0.99 \\
\texttt{AE-KD} & 0.68 & 0.69 & 0.69 & 0.96 & 0.96 & 0.96 & 0.55 & 0.61 & 0.82 & 0.88 & 0.93 & 0.99 & 0.27 & 0.69 & 0.73 & 0.62 & 0.78 & 0.94 \\
\texttt{Reinforced} & 0.69 & 0.70 & 0.71 & 0.96 & 0.96 & 0.96 & 0.79 & 0.81 & 0.82 & 0.99 & 0.99 & 0.99 & 0.18 & 0.24 & 0.28 & 0.54 & 0.65 & 0.74 \\
\texttt{CAWPE} & 0.68 & 0.69 & 0.70 & 0.96 & 0.96 & 0.96 & 0.74 & 0.78 & 0.81 & 0.99 & 0.99 & 0.99 & 0.24 & 0.42 & 0.43 & 0.40 & 0.57 & 0.76 \\
\hline
\texttt{AED-LOO} & 0.71 & 0.72 & \textbf{0.73} & 0.96 & \textbf{0.97} & 0.96 & \textbf{0.94} & \textbf{0.95} & \textbf{0.95} & \textbf{1.00} & \textbf{1.00} & \textbf{1.00} & 0.96 & 0.98 & 0.98 & \textbf{1.00} & \textbf{1.00} & \textbf{1.00} \\
\texttt{LightTS} & \textbf{0.72} & \textbf{0.73} & \textbf{0.73} & \textbf{0.97} & \textbf{0.97} & \textbf{0.97} & \textbf{0.94} & 0.94 & \textbf{0.95} & \textbf{1.00} & \textbf{1.00} & \textbf{1.00} & \textbf{0.99} & \textbf{1.00} & \textbf{1.00} & \textbf{1.00} & \textbf{1.00} & \textbf{1.00} \\
\hline
\textbf{} & \multicolumn{6}{l|}{\textit{FaceAll}} & \multicolumn{6}{l|}{\textit{Phoneme}} & \multicolumn{6}{l|}{\textit{UWave}}\\
\hline
\texttt{FP-Ensem/FP-Stud} & \multicolumn{3}{c|}{0.81 / \boxed{0.85}} & \multicolumn{3}{c|}{0.97 / \boxed{0.99}} & \multicolumn{3}{c|}{0.29 / \boxed{0.30}} & \multicolumn{3}{c|}{0.61 / \boxed{0.64}} & \multicolumn{3}{c|}{0.93 / \boxed{0.95}} & \multicolumn{3}{c|}{1.00 / \boxed{1.00}} \\
\hhline{|~|-|-|-|-|-|-|-|-|-|-|-|-|-|-|-|-|-|-|}
\texttt{Classic KD} & 0.75 & 0.77 & 0.80 & 0.84 & 0.88 & 0.89 & 0.26 & 0.27 & \textbf{0.29} & 0.59 & 0.61 & 0.62 & 0.75 & 0.79 & 0.81 & \textbf{1.00} & \textbf{1.00} & \textbf{1.00} \\
\texttt{AE-KD} & 0.74 & 0.76 & 0.75 & 0.83 & 0.88 & 0.94 & 0.24 & 0.23 & 0.26 & 0.56 & 0.57 & 0.58 & 0.77 & 0.79 & 0.81 & \textbf{1.00} & \textbf{1.00} & \textbf{1.00} \\
\texttt{Reinforced} & 0.68 & 0.72 & 0.79 & 0.85 & 0.86 & 0.86 & 0.26 & \textbf{0.28} & \textbf{0.29} & 0.60 & 0.60 & 0.62 & 0.72 & 0.72 & 0.80 & \textbf{1.00} & 0.99 & 0.99 \\
\texttt{CAWPE} & 0.69 & 0.70 & 0.76 & 0.83 & 0.85 & 0.86 & 0.22 & 0.23 & 0.26 & 0.58 & 0.61 & \textbf{0.63} & 0.66 & 0.79 & 0.81 & \textbf{1.00} & 0.99 & 0.99 \\
\hline
\texttt{AED-LOO} & \textbf{0.81} & \textbf{0.83} & 0.82 & \textbf{0.99} & \textbf{0.98} & \textbf{0.98} & \textbf{0.27} & 0.26 & 0.26 & \textbf{0.61} & 0.60 & 0.60 & 0.86 & 0.87 & 0.87 & \textbf{1.00} & \textbf{1.00} & \textbf{1.00} \\
\texttt{LightTS} & \textbf{0.81} & 0.82 & \textbf{0.84} & 0.98 & \textbf{0.98} & \textbf{0.98} & 0.26 & \textbf{0.27} & \textbf{0.27} & \textbf{0.61} & \textbf{0.62} & \textbf{0.63} & \textbf{0.87} & \textbf{0.88} & \textbf{0.89} & \textbf{1.00} & \textbf{1.00} & \textbf{1.00} \\
\hline
    \end{tabular}
\end{table*}

The main observation is the leading performance in terms of {\textit{Accuracy}} and \textit{Top-5 Accuracy} of \texttt{LightTS} on all data sets. The metrics for \texttt{LightTS} are consistently better on all data sets, and they are close to the accuracy of \texttt{FP-Ensem}. 
On data sets such as \textit{Adiac} and \textit{PigArt}, some of the quantized models are able to outperform \texttt{FP-Ensem}, which can be explained by the two sources of training in the quantized student, i.e., distillation loss and the classification loss.
However, even when quantized models outperform the ensemble, they still perform below the upper-bound performance of \texttt{FP-Stud}. 
The condition is explained by the fact that 32-bit provides more information, so that model captures better the complete parameters representation, without any precision loss. Therefore, it is expected that it outperforms all quantization settings since they lose some level of precision.

The \texttt{AED-LOO} baseline achieves very good performance for {\textit{Accuracy}} and \textit{Top-5 Accuracy}. In most of the cases, it is second best. 
This is expected since \texttt{AED-LOO} shares the \texttt{AED} component with \texttt{LightTS}.
Also, the differences between the two methods are related to how they remove teachers. \texttt{LightTS} can evaluate all the $N$ possible ensembles, regardless the changes in accuracy when some teachers are removed, while \texttt{AED-LOO} stops when the accuracy drops because of its factorial growth in the number of evaluations.

The rather reduced performance of the remaining baselines is related to the premises on which they are built. In all cases, they estimate the weights for the classes probabilities and combine them in a single ensemble, instead of considering independently each base model contribution to the student. This works well in full precision models, but it conflicts with the quantization because the final ensemble does not provide enough support for the compressed student in contrast to the independent distillation for each member.

In addition, the evaluation on the \textit{UWave} shows the model applicability in multi\Hyphdash dimensional time series, so the compression process it is not constrained by the input dimensions. Also, the small number of labels in the data set, only eight,
gives the perfect score for the \textit{Top-5 Accuracy} metric for all the baselines.

\noindent
\textbf{Teacher removal:} 
To assess the effect of using the Gumbel-Softmax based teacher removal strategy, we compare it with two variants: using the Softmax function to remove teachers and no removing teachers from the ensemble.
The results are shown in Table~\ref{table:gumble} for the \textit{Adiac} data set. Other data sets show similar results.  
Using the Gumbel-Softmax yields 16--24\% improvement in the \textit{Accuracy} and around 5--13\% improvement in the \textit{Top-5 Accuracy} with respect to Softmax removal and no removal.
Thus, the choice of the Gumbel-Softmax for disregarding base models is justified. 
\begin{table}[H]
    \small
    \centering
    \caption{Teacher Removal Strategies for \textit{Adiac} Data Set.}
    \label{table:gumble}
    \begin{tabular}{ |l|c|c|c|c|c|c| } 
    \hline
    \textbf{} & \multicolumn{3}{c|}{\textit{Accuracy}} & \multicolumn{3}{c|}{\textit{Top-5 Accuracy}}  \\ 
    \hline
    \textbf{Bit-width} & \textit{4-bit} & \textit{8-bit} & \textit{16-bit} & \textit{4-bit} & \textit{8-bit} & \textit{16-bit} \\
    \hline
    \texttt{No removal} & 0.55 & 0.61 & 0.64 & 0.86 & 0.92 & 0.93 \\
    \texttt{Softmax} & 0.53 & 0.57 & 0.63 & 0.84 & 0.85 & 0.92  \\
    \texttt{Gumble} & \textbf{0.77} & \textbf{0.77} & \textbf{0.79} & \textbf{0.97} & \textbf{0.97} & \textbf{0.97} \\
    \hline
    \end{tabular}
\end{table}

\noindent
\textbf{Running time: }
We cover training time and inference time separately. 
%
The training time is evaluated using the complete UCR archive in Figure~\ref{subfig:rankruntime} for all bit settings also considering the null-hypothesis Friedman and the Wilcoxon-Holm post-hoc tests. The ranking places \texttt{Classic KD} and \texttt{AED-One} in the first group. This is expected, as they share the same complexity as shown by the complexity analysis in Section~\ref{sssec:learningAED}. 
\texttt{Reinforced} and \texttt{LightTS} are second, while \texttt{CAWPE} and \texttt{AE-KD} are third, and \texttt{AED-LOO} is last, as it uses an inefficient leave-one-out strategy to remove teachers. 
The overall results for \texttt{LightTS} show an appropriate trade-off between accuracy and training time. It achieves the best accuracy with a very competitive running time, and the efficiency contenders (\texttt{Reinforced}, \texttt{AED-One}, and \texttt{Classic KD}) are in the bottom segment in terms of overall accuracy. 
In addition, the training time on the full UCR archive is 
shown in Figure~\ref{subfig:boxplot} using box plots. 
%
%
%
Next, inference occurs online, where we use the quantized model to perform classification. Its running time depends only on its size (e.g., 4, 8, or 16 bits) and is independent of the distillation method used. 
To conclude, since training occurs offline and thus is often not time critical, \texttt{LightTS} is the best choice as it offers the best accuracy. 
If the training time is critical, \texttt{AED-One} is the best choice as it is as efficient as \texttt{Classic KD}, but is more accurate. 

\input{figures/running_boxplot}

\noindent
\textbf{Non-deep Learning Based Models:}
To verify that \texttt{LightTS} is a generic framework, which can also apply to base models other than \texttt{InceptionTime}, we consider non-deep learning based models, \textit{TDE}, \textit{CIF}, and \textit{Forest}, Table~\ref{table:other_teachers} shows results on data sets \texttt{Adiac} and \texttt{PigAirway}, due to the space limitation.
Results on the remaining data sets exhibit similar trends.



\addtolength{\tabcolsep}{-1pt}
\begin{table*}[ht!]
\setlength\fboxsep{1pt}
    \small
    \centering
    \caption{Accuracy of Lightweight Students, using Ensembles with Base Models other than \texttt{InceptionTime}. }
    \label{table:other_teachers}
    \begin{tabular}{ |l|l|c|c|c|c|c|c|c|c|c|c|c|c|c|c|c|c|c|c| } 
    \hline
&    \textbf{} & \multicolumn{6}{c|}{\textit{TDE}} & \multicolumn{6}{c|}{\textit{CIF}} & \multicolumn{6}{c|}{\textit{Forest}} \\
\hline
   & \textbf{} & \multicolumn{3}{c|}{\textit{Accuracy}} & \multicolumn{3}{c|}{\textit{Top-5 Accuracy}} & \multicolumn{3}{c|}{\textit{Accuracy}} & \multicolumn{3}{c|}{\textit{Top-5 Accuracy}} & \multicolumn{3}{c|}{\textit{Accuracy}} & \multicolumn{3}{c|}{\textit{Top-5 Accuracy}} \\ 
    \hline
   & \textbf{Bit-width} & \textit{4} & \textit{8} & \textit{16} & \textit{4} & \textit{8} & \textit{16}
    & \textit{4} & \textit{8} & \textit{16} & \textit{4} & \textit{8} & \textit{16}
     & \textit{4} & \textit{8} & \textit{16} & \textit{4} & \textit{8} & \textit{16}
    \\ 
    \hline
\parbox[t]{2mm}{\multirow{6}{*}{\rotatebox[origin=c]{90}{\textit{Adiac}}}} &
\texttt{FP-Ensem/FP-Stud} & \multicolumn{3}{c|}{0.50 / \boxed{0.61}} & \multicolumn{3}{c|}{0.79 / \boxed{0.83}} & \multicolumn{3}{c|}{0.79 / \boxed{0.84}} & \multicolumn{3}{c|}{0.97 / \boxed{0.97}} & \multicolumn{3}{c|}{0.74 / \boxed{0.77}} & \multicolumn{3}{c|}{0.95 / \boxed{0.96}}\\
\hhline{|~|-|-|-|-|-|-|-|-|-|-|-|-|-|-|-|-|-|-|-|}
& \texttt{Classic KD} & 0.14 & 0.13 & 0.12 & 0.37 & 0.32 & 0.31 & 0.24 & 0.27 & 0.24 & 0.48 & 0.43 & 0.56 & 0.21 & 0.23 & 0.18 & 0.57 & 0.42 & 0.40 \\
& \texttt{AE-KD} & 0.13 & 0.18 & 0.16 & 0.35 & 0.41 & 0.37 & 0.21 & 0.24 & 0.21 & 0.46 & 0.43 & 0.47 & 0.23 & 0.29 & 0.21 & 0.55 & 0.71 & 0.49 \\
& \texttt{Reinforced} & 0.11 & 0.13 & 0.14 & 0.30 & 0.31 & 0.30 & 0.24 & 0.19 & 0.19 & 0.53 & 0.37 & 0.37 & 0.15 & 0.18 & 0.19 & 0.41 & 0.36 & 0.39 \\
& \texttt{CAWPE} & 0.12 & 0.15 & 0.14 & 0.34 & 0.35 & 0.33 & 0.15 & 0.14 & 0.14 & 0.35 & 0.31 & 0.34 & 0.12 & 0.15 & 0.14 & 0.34 & 0.35 & 0.33 \\
& \texttt{LightTS} & \textbf{0.32} & \textbf{0.37} & \textbf{0.41} & \textbf{0.65} & \textbf{0.69} & \textbf{0.75} & \textbf{0.66} & \textbf{0.71} & \textbf{0.71} & \textbf{0.93} & \textbf{0.94} & \textbf{0.95} & \textbf{0.68} & \textbf{0.69} & \textbf{0.71} & \textbf{0.93} & \textbf{0.93} & \textbf{0.95} \\

\hline
\parbox[t]{2mm}{\multirow{6}{*}{\rotatebox[origin=c]{90}{\textit{PigAirway}}}} &
\texttt{FP-Ensem/FP-Stud} & \multicolumn{3}{c|}{0.76 / \boxed{0.80}} & \multicolumn{3}{c|}{0.97 / \boxed{0.98}} & \multicolumn{3}{c|}{0.42 / \boxed{0.46}} & \multicolumn{3}{c|}{0.93 / \boxed{0.94}} & \multicolumn{3}{c|}{0.42 / \boxed{0.45}} & \multicolumn{3}{c|}{0.94 / \boxed{0.95}} \\
\hhline{|~|-|-|-|-|-|-|-|-|-|-|-|-|-|-|-|-|-|-|-|}
& \texttt{Classic KD} & 0.22 & 0.23 & 0.14 & 0.63 & 0.63 & 0.42 & 0.11 & 0.13 & 0.11 & 0.33 & 0.39 & 0.34 & 0.14 & 0.15 & 0.10 & 0.40 & 0.43 & 0.34 \\
& \texttt{AE-KD} & 0.15 & 0.20 & 0.18 & 0.39 & 0.49 & 0.49 & 0.11 & 0.13 & 0.17 & 0.35 & 0.35 & 0.47 & 0.11 & 0.10 & 0.11 & 0.33 & 0.33 & 0.32 \\
& \texttt{Reinforced} & 0.13 & 0.13 & 0.19 & 0.36 & 0.43 & 0.49 & 0.12 & 0.10 & 0.07 & 0.29 & 0.31 & 0.27 & 0.08 & 0.10 & 0.14 & 0.25 & 0.37 & 0.40 \\
& \texttt{CAWPE} & 0.11 & 0.12 & 0.12 & 0.33 & 0.38 & 0.35 & 0.11 & 0.12 & 0.12 & 0.32 & 0.35 & 0.36 & 0.11 & 0.12 & 0.12 & 0.33 & 0.38 & 0.35 \\
& \texttt{LightTS} & \textbf{0.60} & \textbf{0.66} & \textbf{0.70} & \textbf{0.93} & \textbf{0.95} & \textbf{0.96} & \textbf{0.36} & \textbf{0.40} & \textbf{0.45} & \textbf{0.68} & \textbf{0.73} & \textbf{0.77} & \textbf{0.33} & \textbf{0.38} & \textbf{0.42} & \textbf{0.66} & \textbf{0.70} & \textbf{0.74} \\
\hline
    \end{tabular}
\end{table*}

The main observation is that \texttt{LightTS} offers much better \textit{Accuracy} and \textit{Top-5 Accuracy} performance than the baselines.
Specifically for \textit{Accuracy}, the level of improvement with respect to the baselines is, for most of the cases, around a factor of three.
The reason for the large difference is the adaptability of \texttt{LightTS} when choosing teachers.
It is able to identify the candidates that are better aligned with the student settings and distills knowledge from them, while the other methods include all the base models regardless of their performance.

Next, in all cases, the full-precision ensemble outperforms the distilled cases at \textit{Accuracy}, meaning that there is still room for improvement of the quantized student models.
As the differences between the non-deep learning base models and \texttt{LightTS} are higher than when using \texttt{InceptionTime} as base models, it appears that the architectural differences between the teachers and the student have some effect on the distillation performance.

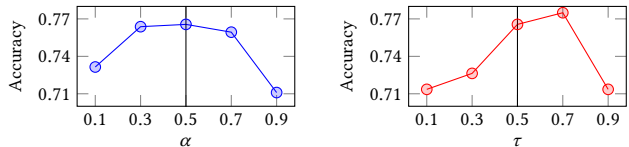
\begin{figure}[t]
\footnotesize
\centering
\begin{subfigure}{0.48\linewidth}
    \begin{tikzpicture}
        \begin{axis}[
            xlabel=$\alpha$,
            ylabel=Accuracy,
            ymax = 0.78,
            ymin = 0.70,
            xtick = {0.1,0.3,0.5,0.7,0.9},
            ytick = {0.71,0.74,0.77},
            width=1.1\linewidth,
            height=0.40*\axisdefaultheight,
            legend pos= outer north east]
            \addplot[blue,fill=none,mark=*, fill opacity=0.2] table[x=alpha, y=accuracy] {figures/data/Alpha.txt};
            \addplot [black, mark=none] coordinates {(0.5, 0.68) (0.5, 0.79)};
        \end{axis}
    \end{tikzpicture}
    \label{subfig:alpha}
\end{subfigure}
\hfill
\begin{subfigure}{0.48\linewidth}
    \begin{tikzpicture}
        \begin{axis}[
            xlabel=$\tau$,
            ylabel=Accuracy,
            ymax = 0.78,
            ymin = 0.70,
            xtick = {0.1,0.3,0.5,0.7,0.9},
            ytick = {0.71,0.74,0.77},
            width=1.1\linewidth,
            height=0.40*\axisdefaultheight,
            legend pos= outer north east]
            \addplot[red,fill=none,mark=*, fill opacity=0.2] table[x=tau, y=accuracy] {figures/data/Tau.txt};
            \addplot [black, mark=none] coordinates {(0.5, 0.68) (0.5, 0.79)};
        \end{axis}
    \end{tikzpicture}
    \label{subfig:tau}
\end{subfigure}
\caption{Hyper-parameter Sensitivity, \texttt{Adiac}, 4-bit.}
\label{fig:sensitivity}
\end{figure}

\noindent
\textbf{Hyper-parameter sensitivity:}
In Figure~\ref{fig:sensitivity}, the sensitivity for the hyper-parameters $\alpha$ and $\tau$ is evaluated for the \texttt{Adiac} data set in a configuration of 4 bits. It shows that $\alpha$ is stable when the two losses are balanced. Then, the accuracy changes are steeper with $\tau$ modifications, meaning that it leads to a different selection of teachers, affecting the overall performance. The choice of $0.5$ for both parameters seems precise, since it is among the best possible options. Other data sets and bit settings show similar trends. 

\begin{figure}[t]
\small
\centering
\hspace*{-5ex}
\begin{subfigure}{0.33\linewidth}
\vspace*{2.3em}
    \begin{tikzpicture}
        \begin{axis}[
            xlabel=$N$,
            ylabel=Accuracy,
            axis y line*=left,
            ylabel shift = -4 pt,
            ylabel style={font=\footnotesize},
            tick label style={font=\footnotesize},
            width=1\linewidth,
            height=0.40*\axisdefaultheight,
            legend pos= outer north east]
            \addplot[blue,fill=none,mark=*, fill opacity=0.2] table[x=base, y=accuracy] {figures/data/scalability_Adiac.txt};
        \end{axis}
        \begin{axis}[ybar,bar width=4pt,
            ylabel=Time (s),
            axis y line*=right,
            axis x line=none,
            ylabel shift = -4 pt,
            ylabel style={font=\footnotesize},
            tick label style={font=\footnotesize},
            width=1\linewidth,
            height=0.40*\axisdefaultheight,
            legend pos= outer north east]
            \addplot[purple,fill=purple,mark=none,fill opacity=0.2] table[x=base, y=running] {figures/data/scalability_Adiac.txt};
        \end{axis}
    \end{tikzpicture}
    \label{subfig:adiac}
    \vspace*{-1.5em}
    \caption{Adiac.}
\end{subfigure}
\hspace*{1ex}
\begin{subfigure}{0.33\linewidth}
    \begin{tikzpicture}
        \begin{axis}[
            xlabel=$N$,
            ylabel=Accuracy,
            axis y line*=left,
            ylabel shift = -4 pt,
            ylabel style={font=\footnotesize},
            tick label style={font=\footnotesize},
            width=1\linewidth,
            height=0.4*\axisdefaultheight,
            legend pos= outer north east]
            \addplot[blue,fill=none,mark=*, fill opacity=0.2] table[x=base, y=accuracy] {figures/data/scalability_PigAirway.txt}; \label{scale:accuracy}
            \label{accuracy_plot}
        \end{axis}
        \begin{axis}[ybar,bar width=4pt,
            ylabel=Time (s),
            axis y line*=right,
            axis x line=none,
            ylabel shift = -4 pt,
            ylabel style={font=\footnotesize},
            tick label style={font=\footnotesize},
            width=1\linewidth,
            height=0.4*\axisdefaultheight,
            legend style={at={(0,1.6)},                   anchor=north, legend columns=2}]            
           \addplot[purple,fill=purple,mark=none, fill opacity=0.2] table[x=base, y=running] {figures/data/scalability_PigAirway.txt}; \label{scale:time}
        \end{axis}
        \matrix [matrix of nodes,nodes={anchor=north},fill=white,draw,inner sep=1pt,row sep=1pt,
        xshift=20pt,yshift=50pt,font=\footnotesize] {
        \ref{scale:accuracy} Accuracy 
        \ref{scale:time} Time \\
        };
    \end{tikzpicture}
    \label{subfig:pigairway}
    \vspace*{-1.5em}
    \caption{PigAirway.}
\end{subfigure}
\begin{subfigure}{0.33\linewidth}
\vspace*{2.3em}
    \begin{tikzpicture}
        \begin{axis}[
            xlabel=$N$,
            ylabel=Accuracy,
            axis y line*=left,
            ylabel shift = -4 pt,
            ylabel style={font=\footnotesize},
            tick label style={font=\footnotesize},
            width=1\linewidth,
            height=0.4*\axisdefaultheight,
            legend pos= outer north east]
            \addplot[blue,fill=none,mark=*, fill opacity=0.2] table[x=base, y=accuracy]{figures/data/scalability_NonInv2.txt};
        \end{axis}
        \begin{axis}[ybar,bar width=4pt,
            ylabel=Time (s),
            axis y line*=right,
            axis x line=none,
            ylabel shift = -4 pt,
            ylabel style={font=\footnotesize},
            tick label style={font=\footnotesize},
            scaled y ticks=base 10:-3,
            ytick scale label code/.code={},
            yticklabel={\pgfmathprintnumber{\tick} K},
            width=1\linewidth,
            height=0.4*\axisdefaultheight]
            \addplot[purple,fill=purple,mark=none,fill opacity=0.2] table[x=base, y=running] {figures/data/scalability_NonInv2.txt};
        \end{axis}
    \end{tikzpicture}
    \label{subfig:noninv2}
    \vspace*{-1.5em}
    \caption{NonInvECG2.}
\end{subfigure}
\hspace*{-5ex}
\caption{Effect of the Number of Base Models $N$.}
\label{fig:scalability}
\end{figure}
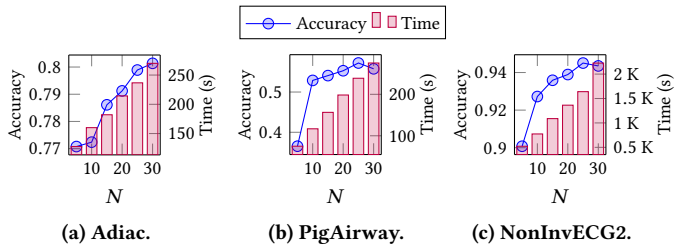

\noindent
\textbf{Effect of the number of base models $\mathbf{N}$:}
Figure~\ref{fig:scalability} shows the impact of the number of base models $N$ on \textit{Accuracy} and \textit{Training time} for data sets \textit{Adiac}, \textit{PigAirway}, and \textit{NonInvECCG2}. In all cases, when $N$ is small, the accuracy results are significantly affected mainly because 
the opportunities 
to choose appropriate base models and to disregard unsuitable base models are reduced. Then, as $N$ increases, the accuracy is relatively stable within a range of 5\%, achieving the 
best accuracy with some 25 or 30 base models.
For \textit{PigAirway} and \textit{NonInvECCG2}, the results decrease slightly when the number of base models reaches 30, which we attribute to the added uncertainty caused by having more base models, i.e., removing base models becomes increasingly arbitrary as more options are available. The changes for \textit{Adiac} are very slightly after considering 25 models, suggesting it will reach a maximum closer closer to that range. The total training time grows linearly as the number of base models increases, which is consistent with the complexity analysis.

\subsubsection{Problem Scenario 2}

To find optimal student settings using the proposed encoded multi-objective Bayesian optimization, we initialize the process with $P=10$ random settings and search for 40 additional settings, meaning that $Q=50$.
We show the results on the data set \texttt{Adiac} given the space limitation. 
Experiments conducted in other data sets show similar observations.

\noindent
\textbf{Feature transformation effect: } 
To evaluate the effect of using the two-phase encoder, we assess it in comparison to (1) a single phase encoder using the reconstruction loss only (i.e., an autoencoder) but without considering the accuracy adjustment loss function, (2) using the original space where the values are normalized, and (3) the original discrete space.
We use these different spaces to train a GP accuracy estimator and evaluate the estimated accuracy vs. ground truth accuracy of 50 randomly sampled settings. 
The results for data sets \textit{Adiac}, \textit{PigAirway}, and \textit{NonInvECCG2} in Table~\ref{table:estimation_gp} show that 
the proposed two-phase encoder gives the most accurate estimation for the unevaluated settings' accuracy, which outperforms the one-phase autoencoder. In addition, simply normalizing the values in settings fails to improve accuracy.

\begin{table}[H]
    \small
    \centering
    \caption{Gaussian Processes Accuracy Estimation.}
    \label{table:estimation_gp}
    \begin{tabular}{ |l|c|c|c|c|c|c| } 
    \hline
    & \multicolumn{2}{l|}{\textit{Adiac}} & \multicolumn{2}{l|}{\textit{PigAirway}} & \multicolumn{2}{l|}{\textit{NonInvECG2}}\\
    \hline
    \textbf{} & \textit{MAE} & \textit{MAPE} & \textit{MAE} & \textit{MAPE} & \textit{MAE} & \textit{MAPE}  \\ 
    \hline
\texttt{Original} & 0.12 & 0.31 & 0.05 & 0.25 & \textbf{0.04} & 0.06 \\
\texttt{Normalized}  & 0.11  & 0.31 & 0.05 & 0.27 & 0.05 & 0.06 \\
\texttt{Single Encoder}  & 0.09 & 0.24 & 0.05 & 0.23 & \textbf{0.04} & \textbf{0.05} \\
\texttt{Two-phase Encoder} & \textbf{0.08} & \textbf{0.23} & \textbf{0.04} & \textbf{0.16} & \textbf{0.04} & \textbf{0.05} \\
\hline
    \end{tabular}
\end{table}

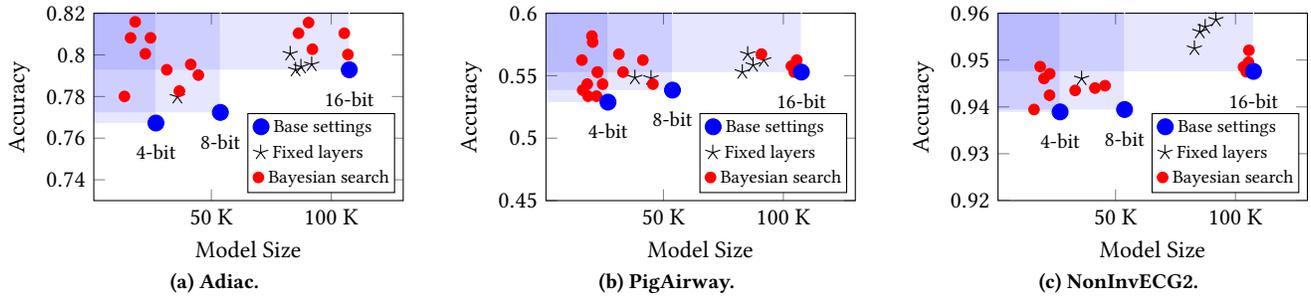
\begin{figure*}
\centering
\begin{subfigure}{0.32\textwidth}
    \edef\windows{"","","","","","","","","","","","","4-bit","","","","","","","","","","","","","","8-bit","","","","","","","","","","","16-bit"}
    \begin{tikzpicture}
        \begin{axis}[
            xlabel=Model Size,
            ylabel=Accuracy,
            ymin=0.73,
            ymax=0.82,
            xmin=1000,
            xmax=130000,
            scaled x ticks=base 10:-3,
            xtick scale label code/.code={},
            xticklabel={\pgfmathprintnumber{\tick} K},
            width=1*\linewidth,
            height=0.56*\axisdefaultheight,
            legend pos= south east]
            \addplot[black, only marks,mark=star,mark size=3pt] table[x=Cost, y=Fixed] {figures/data/Adiac_BO.txt}; \label{fixed}
            \addplot[red, only marks] table[x=Cost, y=Optimized] {figures/data/Adiac_BO.txt}; \label{opt}
            \addplot[blue,mark size=3pt, only marks,mark=*, nodes near coords=\pgfmathsetmacro{\winstring}{{\windows}[\coordindex]}\winstring, nodes near coords style={text=black,font=\small,xshift=0pt,yshift=-5pt,anchor=north}] table[x=Cost, y=Base] {figures/data/Adiac_BO.txt}; \label{base}
            \addplot [color=white,fill=blue, fill opacity=0.1]coordinates {(0, 0.767263427) (0, 1) (26880, 1) (26880, 0.767263427) };
            \addplot [color=white,fill=blue, fill opacity=0.1]coordinates {(0, 0.772378517) (0, 1) (53760, 1) (53760, 0.772378517) };
            \addplot [color=white,fill=blue, fill opacity=0.1]coordinates {(0, 0.792838875) (0, 1) (107520, 1) (107520, 0.792838875) };
        \end{axis}
        \matrix [matrix of nodes,nodes={anchor=west},fill=white,draw,inner sep=1pt,row sep=1pt,
        xshift=86pt,yshift=18pt,font=\footnotesize] {
        \ref{base} Base settings \\
        \ref{fixed} Fixed layers \\
        \ref{opt} Bayesian search \\
        };
    \end{tikzpicture}
    \vspace*{-0.5em}
    \caption{Adiac.}
    \label{fig:adiac_bo}
\end{subfigure}
\hfill
\begin{subfigure}{0.32\textwidth}
    \edef\windows{"","","","","","","","","","","","","","","","4-bit","","","","","","","","","","","","8-bit","","","","","","","","","","","","","","16-bit"}
    \begin{tikzpicture}
        \begin{axis}[
            xlabel=Model Size,
            ylabel=Accuracy,
            ymin=0.45,
            ymax=0.60,
            xmin=1000,
            xmax=130000,
            scaled x ticks=base 10:-3,
            xtick scale label code/.code={},
            xticklabel={\pgfmathprintnumber{\tick} K},
            width=1*\linewidth,
            height=0.56*\axisdefaultheight,
            legend pos= south east]
            \addplot[black, only marks,mark=star,mark size=3pt] table[x=Cost, y=Fixed] {figures/data/PigAir_BO.txt};
            \addplot[red, only marks] table[x=Cost, y=Optimized] {figures/data/PigAir_BO.txt};
            \addplot[blue,mark size=3pt, only marks,mark=*, nodes near coords=\pgfmathsetmacro{\winstring}{{\windows}[\coordindex]}\winstring, nodes near coords style={text=black,font=\small,xshift=0pt,yshift=-5pt,anchor=north}] table[x=Cost, y=Base] {figures/data/PigAir_BO.txt};
            \addplot [color=white,fill=blue, fill opacity=0.1]coordinates {(0, 0.528846154) (0, 1) (26880, 1) (26880, 0.528846154) };
            \addplot [color=white,fill=blue, fill opacity=0.1]coordinates {(0, 0.538461538) (0, 1) (53760, 1) (53760, 0.538461538) };
            \addplot [color=white,fill=blue, fill opacity=0.1]coordinates {(0, 0.552884615) (0, 1) (107520, 1) (107520, 0.552884615) };
        \end{axis}
        \matrix [matrix of nodes,nodes={anchor=west},fill=white,draw,inner sep=1pt,row sep=1pt,
        xshift=86pt,yshift=18pt,font=\footnotesize] {
        \ref{base} Base settings \\
        \ref{fixed} Fixed layers \\
        \ref{opt} Bayesian search \\
        };
    \end{tikzpicture}
    \vspace*{-0.5em}
    \caption{PigAirway.}
    \label{fig:pigair_bo}
\end{subfigure}
\hfill
\begin{subfigure}{0.32\textwidth}
    \edef\windows{"","","","","","","","","","","4-bit","","","","","8-bit","","","","","","","","","","16-bit"}
    \begin{tikzpicture}
        \begin{axis}[
            xlabel=Model Size,
            ylabel=Accuracy,
            ymin=0.92,
            ymax=0.96,
            xmin=1000,
            xmax=130000,
            scaled x ticks=base 10:-3,
            xtick scale label code/.code={},
            xticklabel={\pgfmathprintnumber{\tick} K},
            width=1*\linewidth,
            height=0.56*\axisdefaultheight,
            legend pos= south east]
            \addplot[black, only marks,mark=star,mark size=3pt] table[x=Cost, y=Fixed] {figures/data/Non2_BO.txt};
            \addplot[red, only marks] table[x=Cost, y=Optimized] {figures/data/Non2_BO.txt};
            \addplot[blue,mark size=3pt, only marks,mark=*, nodes near coords=\pgfmathsetmacro{\winstring}{{\windows}[\coordindex]}\winstring, nodes near coords style={text=black,font=\small,xshift=0pt,yshift=-5pt,anchor=north}] table[x=Cost, y=Base] {figures/data/Non2_BO.txt};
            \addplot [color=white,fill=blue, fill opacity=0.1]coordinates {(0, 0.938931298) (0, 1) (26880, 1) (26880, 0.938931298) };
            \addplot [color=white,fill=blue, fill opacity=0.1]coordinates {(0, 0.939440204) (0, 1) (53760, 1) (53760, 0.939440204) };
            \addplot [color=white,fill=blue, fill opacity=0.1]coordinates {(0, 0.947582697) (0, 1) (107520, 1) (107520, 0.947582697) };
        \end{axis}
        \matrix [matrix of nodes,nodes={anchor=west},fill=white,draw,inner sep=1pt,row sep=1pt,
        xshift=86pt,yshift=18pt,font=\footnotesize] {
        \ref{base} Base settings \\
        \ref{fixed} Fixed layers \\
        \ref{opt} Bayesian search \\
        };
    \end{tikzpicture}
    \vspace*{-0.5em}
    \caption{NonInvECG2.}
    \label{fig:non2_bo}
\end{subfigure}
\vspace*{-1em}
\caption{Base Settings Improvement.}
\label{fig:bo_all}
\vspace*{-1em}
\end{figure*}
\begin{figure*}
\centering
\begin{subfigure}{0.32\textwidth}
    \begin{tikzpicture}
        \begin{axis}[
            xlabel=Model Size,
            ylabel=Accuracy,
            ymin=0.67,
            ymax=0.82,
            xmin=11000,
            xmax=48500,
            scaled x ticks=base 10:-3,
            xtick scale label code/.code={},
            xticklabel={\pgfmathprintnumber{\tick} K},
            ytick = {0.69,0.73,0.77,0.81},
            width=1*\linewidth,
            height=0.5*\axisdefaultheight]
            \addplot[black,fill=gray, mark=square, fill opacity=0.1] table[x=Cost, y=Value] {figures/data/Adiac_Multi_E_P.txt}\closedcycle; \label{multi}
            \addplot[color=red,fill=red, mark=*, fill opacity=0.1] table[x=Cost, y=Value] {figures/data/Adiac_Multi_P.txt}\closedcycle; \label{single}
            \addplot[blue,fill=blue, mark=triangle, fill opacity=0.2] table[x=Cost, y=Value] {figures/data/Adiac_Random_P.txt}\closedcycle; \label{random}
        \end{axis}
        \matrix [matrix of nodes,nodes={anchor=west},fill=white,draw,inner sep=1pt,row sep=1pt,
        xshift=81pt,yshift=13pt,font=\footnotesize] {
        \ref{multi} Encoded MOBO \\
        \ref{single} MOBO \\
        \ref{random} Random \\
        };
    \end{tikzpicture}
    \caption{Adiac.}
    \label{fig:adiac_pareto}
\end{subfigure}
\hfill
\begin{subfigure}{0.32\textwidth}
    \begin{tikzpicture}
        \begin{axis}[
            xlabel=Model Size,
            ylabel=Accuracy,
            ymin=0.34,
            ymax=0.61,
            xmin=11000,
            xmax=48500,
            scaled x ticks=base 10:-3,
            xtick scale label code/.code={},
            xticklabel={\pgfmathprintnumber{\tick} K},
            ytick = {0.36,0.44,0.52,0.60},
            width=1*\linewidth,
            height=0.5*\axisdefaultheight,
            legend pos= south east]
            \addplot[black,fill=gray, mark=square, fill opacity=0.1] table[x=Cost, y=Value] {figures/data/PigAir_Multi_E_P.txt}\closedcycle;
            \addplot[color=red,fill=red, mark=*, fill opacity=0.1] table[x=Cost, y=Value] {figures/data/PigAir_Multi_P.txt}\closedcycle;
            \addplot[blue,fill=blue, mark=triangle, fill opacity=0.2] table[x=Cost, y=Value] {figures/data/PigAir_Random_P.txt}\closedcycle;
        \end{axis}
        \matrix [matrix of nodes,nodes={anchor=west},fill=white,draw,inner sep=1pt,row sep=1pt,
        xshift=81pt,yshift=13pt,font=\footnotesize] {
        \ref{multi} Encoded MOBO \\
        \ref{single} MOBO &\\
        \ref{random} Random \\
        };
    \end{tikzpicture}
    \caption{PigAirway.}
    \label{fig:pigair_pareto}
\end{subfigure}
\hfill
\begin{subfigure}{0.32\textwidth}
    \begin{tikzpicture}
        \begin{axis}[
            xlabel=Model Size,
            ylabel=Accuracy,
            ymin=0.82,
            ymax=0.96,
            xmin=11000,
            xmax=49900,
            scaled x ticks=base 10:-3,
            xtick scale label code/.code={},
            xticklabel={\pgfmathprintnumber{\tick} K},
            ytick = {0.83,0.87,0.91,0.95},
            width=1*\linewidth,
            height=0.5*\axisdefaultheight,
            legend pos= south east]
            \addplot[black,fill=gray, mark=square, fill opacity=0.1] table[x=Cost, y=Value] {figures/data/Non_Multi_E_P.txt}\closedcycle;
            \addplot[color=red,fill=red, mark=*, fill opacity=0.1] table[x=Cost, y=Value] {figures/data/Non_Multi_P.txt}\closedcycle;
            \addplot[blue,fill=blue, mark=triangle, fill opacity=0.2] table[x=Cost, y=Value] {figures/data/Non_Random_P.txt}\closedcycle;
        \end{axis}
        \matrix [matrix of nodes,nodes={anchor=west},fill=white,draw,inner sep=1pt,row sep=1pt,
        xshift=81pt,yshift=13pt,font=\footnotesize] {
        \ref{multi} Encoded MOBO \\
        \ref{single} MOBO \\
        \ref{random} Random \\
        };
    \end{tikzpicture}
    \caption{NonInvECG2.}
    \label{fig:non2_pareto}
\end{subfigure}
\vspace*{-1em}
\caption{Identifying Pareto Frontiers.}
\label{fig:opt_all}
\vspace*{-1.5em}

\end{figure*}
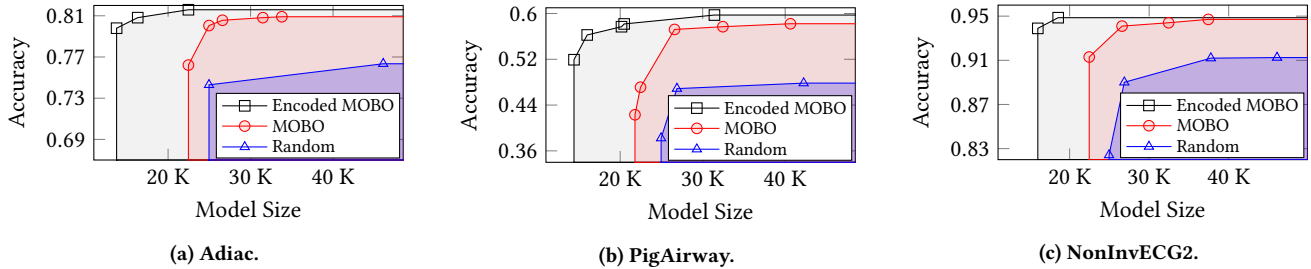
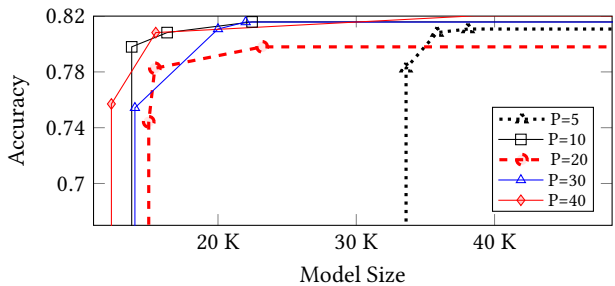
\begin{figure}
    \begin{tikzpicture}
        \begin{axis}[
            xlabel=Model Size,
            ylabel=Accuracy,
            ymin=0.67,
            ymax=0.82,
            xmin=11000,
            xmax=48500,
            ytick = {0.70,0.74,0.78,0.82},
            scaled x ticks=base 10:-3,
            xtick scale label code/.code={},
            xticklabel={\pgfmathprintnumber{\tick} K},
            width=1*\linewidth,
            height=0.6*\axisdefaultheight]
            \addplot[black, very thick,dotted, mark=triangle, fill opacity=0.2] table[x=Cost, y=Value] {figures/data/Adiac_P5_E.txt}; \label{p5}
            \addplot[black, mark=square, fill opacity=0.2] table[x=Cost, y=Value] {figures/data/Adiac_Multi_E_P.txt}; \label{p10}
            \addplot[red, very thick,dashed,mark=*, fill opacity=0.1] table[x=Cost, y=Value] {figures/data/Adiac_P20.txt}; \label{p20}
            \addplot[color=blue, mark=triangle, fill opacity=0.1] table[x=Cost, y=Value] {figures/data/Adiac_P30.txt}; \label{p30}
            \addplot[red, mark=diamond*, fill opacity=0.2] table[x=Cost, y=Value] {figures/data/Adiac_P40.txt}; \label{p40}
        \end{axis}
        \matrix [matrix of nodes,nodes={anchor=west},fill=white,draw,inner sep=1pt,row sep=1pt,
        xshift=170pt,yshift=25pt,font=\footnotesize] {
        \ref{p5} P=5 \\
        \ref{p10} P=10 &\\
        \ref{p20} P=20 &\\
        \ref{p30} P=30 &\\
        \ref{p40} P=40 &\\
        };
    \end{tikzpicture}
    \vspace*{-1em}
    \caption{Varying $P$, Encoded MOBO, Adiac.}
    \vspace*{-2em}
    \label{fig:p_sensitivity}
\end{figure}

\noindent
\textbf{Base settings improvement:}
In Figure~\ref{fig:adiac_bo}, \ref{fig:pigair_bo}, and \ref{fig:non2_bo}, 
we show how the encoded Bayesian optimization is able to improve the results of the fixed settings used in the first set of experiments for the \textit{Adiac}, \textit{PigAirway}, and \textit{NonInvECCG2} data sets, respectively.
In the figure, we include the three lightweight student cases of 4, 8, and 16 bits (c.f. Table~\ref{table:Main}) denoted as ``Base settings,'' and denoting upper-left areas for possible improvements, meaning that any point in that area has a better accuracy at a reduced size.
We only show the results in the areas for improvement to keep the figures clear, so fewer than $Q=50$ points are shown.
Then, we illustrate the effect of the optimization with two levels of flexibility.
First, we keep the number of layers fixed as the base settings and apply the optimization limiting the search space to only bit-width. 
In the figure, these are denoted as ``Fixed layers.''
Then, we explore the complete search space, varying the number of layers, filter length, and bit-width, shown in the figure as ``Encoded MOBO.'' 
The evaluation shows that the optimization process is able to find better student settings than with the base settings, i.e., improving accuracy while consuming less space.
In addition, the figure shows 
that exploring only the bit-width space enables better settings, but that it is potentially more beneficial to explore the complete search space, since there are better settings when also the layers can be varied.

\noindent
\textbf{Pareto frontiers:}
Using the \textit{Adiac}, \textit{PigAirway}, and \textit{NonInvECCG2} data sets, we show in Figures~\ref{fig:adiac_pareto}, \ref{fig:pigair_pareto}, and \ref{fig:non2_pareto} 
a comparison of the Pareto frontiers when using different methods.
We include the proposed encoded Multi-objective Bayesian optimization (``Encoded MOBO''); classic Multi-objective Bayesian optimization (``MOBO''), where the settings are not encoded;  
and randomly choosing all $Q$ settings (``Random''). 
%
%
The Encoded MOBO obtains a better Pareto frontier than do the other two methods, i.e., it is closer to the upper, left corner. 
Thus, the frontier includes settings with relatively small model size and high accuracy, meaning that it is possible to find very competitive settings with very strict memory constraints.

The methods are evaluated in terms of running time, as shown in Table~\ref{table:optimization_time}.
The ``Random'' search is faster since it does not optimize the search, but that leads to a poor Pareto frontier. Next, the differences in running time between the ``MOBO'' and ``Encoded MOBO'' optimizations are relatively small, which is expected because of the similar optimization processes. 
\begin{table}
    \small
    \centering
    \caption{Optimization Running Time (hours).}
    \label{table:optimization_time}
    \begin{tabular}{ |l|c|c|c| } 
    \hline
    \textbf{} &  \textit{Adiac} &  \textit{PigAirway} &  \textit{NonInvECG2}\\ 
    \hhline{|-|-|-|-|}
    \texttt{Random} & 4.61 & 2.72 & 19.65 \\
    \texttt{MOBO} & 5.78 & 3.45 & 28.77 \\
    \texttt{Encoded MOBO} & 5.85 & 3.48 & 29.82 \\
    \hline
    \end{tabular}
\end{table}

To evaluate the effect of initializing the optimization process with evaluated settings sets of different size, i.e., with fixed Q while changing $P$, we build the Pareto frontiers with $P$ equal to 5, 10, 20, 30, and 40---see Figure~\ref{fig:p_sensitivity}. Using $P=5$ constrains the process substantially, and it only identifies models of relatively large size, implying that a poorly-initialized GP may mislead the optimization process of BO. In contrast, using $P=10,20,30$ yields similar Pareto frontiers, and we empirically find that $P=10$ is a good choice since it successfully initializes the exploration similarly to larger values.
\section{Related Work} \label{sec:related_work}

\noindent
\textbf{Time Series Classification:} Time series classification has seen substantial advances, and state-of-the-art methods offer impressive accuracy~\cite{FawazFWIM19, RuizFLMB21}. However, the state-of-the-art methods require significant computing resources, making them inapplicable in resource-limited settings.
Some methods (e.g., \cite{MiddlehurstLFLB21}) use large ensembles, that combine the results from dozens of single models to compute a result, while other methods (e.g.,\cite{DempsterPW20, LubbaSKSFJ19, GogolouTEBP20,DempsterSW21}) require multiple data transformations for exploring time series properties~\cite{HoHP21, BoniolP20}, which requires substantial memory.
Then, deep learning~\cite{WangYO17,FawazLFPSWWIMP20} and non-deep learning~\cite{YangSGLJ20,DengRTM13,MiddlehurstVB19,LubbaSKSFJ19,MiddlehurstLB20} based ensembles are competitive models, so they are considered as inputs for \texttt{LightTS}.

\noindent
\textbf{Knowledge Distillation:} Knowledge distillation has been studied since its introduction~\cite{HintonVD15}, with studies considering single teachers~\cite{PolinoPA18,TianKI20,DaoKSM21,RomeroBKCGB14}, ensembles of diverse teachers~\cite{FukudaSKTCR17,YuanSPLGFJ21,DuYLW00Z20,ZhangJS020}, 
and techniques that aim to improve 
performance via self-distillation~\cite{ZhangBM22} and online student feedback~\cite{ChenMWF020}.
When compressing from an ensemble of diverse teachers into a student, existing studies consider mainly a student with full-precision parameters, while we consider lightweight students, e.g., with 4, 8, or 16-bit quantized parameters. This difference calls for a more flexible distillation strategy and means to identify the Pareto optimal frontier, which is not considered by existing studies. 
Although different methods exist that balance different teachers' contributions to an ensemble~\cite{DuYLW00Z20,YuanSPLGFJ21,LargeLB19}, which then guides the knowledge distillation, no studies consider the removal of teachers. We
propose a novel bi-level optimization modeling that not only assigns appropriate weights to useful teachers, but also facilitates removal of irrelevant teachers, thus achieving more flexible distillation and higher accuracy, as shown in the experiments.

\noindent
\textbf{Pareto frontier:} 
Existing studies do not consider how to build a Pareto frontier, partially because they consider only students with full-precision parameters~\cite{TungM18,TianKI20,YimJBK17}. 
In our setting, it is possible to use different bit-widths in different layers, making it challenging to identify Pareto frontiers efficiently. To contend with this challenge, we propose a novel encoding scheme along with an Encoded
MOBO method to identify a Pareto frontier by only evaluating a small number of student model settings. 
%
%
Skyline querying~\cite{BorzsonyiKS01,DBLP:conf/icde/YangGJKS14} is able to
efficiently identify the Pareto frontier for an input data set $D$ in the form of (accuracy, size) pairs. However, naively constructing $D$ requires evaluating the accuracy of large numbers of student settings, which is prohibitively expensive. The
Encoded MOBO aims at constructing an appropriate input data
set $D$ by evaluating the accuracy of a small number of student settings, upon which any skyline querying algorithm can be applied. 
Thus, skyline querying and Encoded MOBO are orthogonal.
\section{Conclusions and Future Work} \label{sec:conclusions}

This paper proposes \texttt{LightTS}, a novel and flexible framework that extends state-of-the-art time series classification to resource-limited devices. First, \texttt{LightTS} is able to adaptively distil the knowledge from a set of high-capacity, large base models to a lightweight model. In doing this, it employs a bi-level optimization approach to estimate the benefit of each base model during the distillation. Second, an encoded multi-objective Bayesian optimization method is proposed to search for the Pareto optimal settings for lightweight models under varying space constraints. The results of experimental studies offer evidence of the effectiveness of \texttt{LightTS} when used on a variety of real-world time series data sets with different base models.

In future work, it is of interest to explore more flexible distillation paradigms, such as online distillation that does not require pretrained base models. It is also of interest to study how to adapt distillation to streaming settings.

\onecolumn
\begin{multicols}{2}
\bibliographystyle{ACM}
\bibliography{references}

\end{multicols}

\end{document}